\documentclass[lettersize,journal]{IEEEtran}

\usepackage[colorlinks,urlcolor=blue,linkcolor=blue,citecolor=blue]{hyperref}
\usepackage{cite}
\usepackage{hyperref}

\ifCLASSINFOpdf
  \usepackage[pdftex]{graphicx}
  \graphicspath{{./img/}}
\else
\fi
\usepackage{amsmath}
\usepackage{amssymb}
\usepackage{mathrsfs}
\usepackage{algorithm}
\usepackage{algpseudocode}

\usepackage{array}
\usepackage{threeparttable}
\usepackage{tabularx}
\usepackage{tablefootnote}
\usepackage{subfigure}

\usepackage{soul}

\usepackage{url}

\usepackage[acronym]{glossaries}
\glsdisablehyper
\usepackage{booktabs}
\usepackage{multirow}
\usepackage{siunitx}
\usepackage{listings}
\usepackage{xcolor}

\lstdefinestyle{mystyle}{
    language=Python,                     
    basicstyle=\ttfamily\tiny,      
    keywordstyle=\color{magenta},         
    commentstyle=\color{blue},            
    stringstyle=\color{red},              
    stepnumber=1,                         
    numbersep=10pt,                       
    backgroundcolor=\color{lightgray!20}, 
    breaklines=true,                      
    showspaces=false,                     
    showstringspaces=false,               
    tabsize=4,                            
}
\newacronym{AI}{AI}{Artificial Intelligence}
\newacronym{NPU}{NPU}{Neural Processing Unit}
\newacronym{EO}{EO}{Electrical-to-Optical}
\newacronym{OE}{OE}{Optical-to-Electrical}
\newacronym{ADC}{ADC}{Analog-to-Digital Converter}
\newacronym{DAC}{DAC}{Digital-to-Analog Converter}
\newacronym{MZI}{MZI}{Mach-Zehnder Interferometer}
\newacronym{MVM}{MVM}{Matrix-Vector Multiplication}
\newacronym{PLC}{PLC}{Plain Light Conversion}
\newacronym{WDM}{WDM}{Wavelength Division Multiplexing}
\newacronym{CMOS}{CMOS}{Complementary Metal-Oxide-Semiconductor}
\newacronym{GPU}{GPU}{Graphics Processing Unit}
\newacronym{CPU}{CPU}{Central Processing Unit}
\newacronym{ML}{ML}{Machine Learning}
\newacronym{PUM}{PUM}{Photonic Unitary Matrix}
\newacronym{SVD}{SVD}{Singulary Value Decomposition}
\newacronym{ONNX}{ONNX}{Open Neural Network Exchange}
\newacronym{DSE}{DSE}{Design Space Exploration}
\newacronym{NN}{NN}{Neural Network}
\newacronym{FFNN}{FFNN}{Feed-Forward Neural Network}
\newacronym{CNN}{CNN}{Convolutional Neural Network}
\newacronym{RNN}{RNN}{Recurrent Neural Network}
\newacronym{PNN}{PNN}{Photonic Neural Network}
\newacronym{SOTA}{SotA}{State-of-the-Art}
\newacronym{RMS}{RMSProp}{Root Mean Square Propagation}
\newacronym{S-Matrix}{S-Matrix}{Scattering Matrix}
\newacronym{CROW}{CROW}{Coupled Resonator Optical Waveguide}

\begin{document}
\bstctlcite{IEEEexample:BSTcontrol}
%
\title{\textbf{L}ux\textbf{I}A: A \textbf{L}ightweight \textbf{U}nitary matri\textbf{X}-based Framework Built on an \textbf{I}terative \textbf{A}lgorithm for Photonic Neural Network Training}

%
%
\IEEEoverridecommandlockouts

\author{
       Tzamn Melendez Carmona,
       Federico Marchesin,
       Marco P. Abrate,
       Peter Bienstman~\IEEEmembership{Member,~IEEE,},
       Stefano Di Carlo~\IEEEmembership{Senior,~IEEE,}
       Alessandro Savino~\IEEEmembership{Senior,~IEEE}

\thanks{This paper was produced by the IEEE Publication Technology Group. They are in Piscataway, NJ.}
\thanks{Manuscript received April 19, 2021; revised August 16, 2021.}

\thanks{Tzamn Melendez Carmona, Stefano Di Carlo, and Alessandro Savino are with the Department of Control and Computer Engineering, Politecnico di Torino, Turin, Italy (emails:\{tzamn.melendez, alessandro.savino, stefano.dicarlo\}@polito.it)}%
\thanks{Federico Marchesin and  Peter Bienstman are with Photonics Research Group, Ghent University - imec, Ghent, Belgium (emails:  \{federico.marchesin, peter.bienstman\}@ugent.be)}%
\thanks{Marco P. Abrate is with Department of Cell and Developmental Biology, University College London, London, UK (e-mail: marco.abrate@ucl.ac.uk)}%
\thanks{This paper has received funding from: The NEUROPULS project in the European Union’s Horizon Europe research and innovation programme under grant agreement No. 101070238.}%
}
%
%

\markboth{Journal of \LaTeX\ Class Files,~Vol.~14, No.~8, August~2015}%
{Shell \MakeLowercase{\textit{et al.}}: Bare Demo of IEEEtran.cls for IEEE Journals}
%

\IEEEpubid{0000--0000/00\$00.00~\copyright~2015 IEEE}


\maketitle
\begin{abstract}

\glspl{PNN} present promising opportunities for accelerating machine learning by leveraging the unique benefits of photonic circuits. However, current \gls{SOTA} \glspl{PNN} simulation tools face significant scalability challenges when training large-scale \glspl{PNN}, due to the computational demands of transfer matrix calculations, resulting in high memory and time consumption. To overcome these limitations, we introduce the Slicing method, an efficient transfer matrix computation approach compatible with back-propagation. We integrate this method into LuxIA, a unified simulation and training framework. The Slicing method substantially reduces memory usage and execution time, enabling scalable simulation and training of large \glspl{PNN}. Experimental evaluations across various photonic architectures and standard datasets—including MNIST, Digits, and Olivetti Faces—show that LuxIA consistently surpasses existing tools in speed and scalability.
Our results advance the \gls{SOTA} in \gls{PNN} simulation, making it feasible to explore and optimize larger, more complex architectures. By addressing key computational bottlenecks, LuxIA facilitates broader adoption and accelerates innovation in \gls{AI} hardware through photonic technologies. This work paves the way for more efficient and scalable photonic neural network research and development.
\end{abstract}


\begin{IEEEkeywords}
Edge-computing, photonics, simulation tools, neural networks, artificial intelligence.
\end{IEEEkeywords}

\glsresetall

%
\section{Introduction} \label{sec:introduction}

\IEEEPARstart{T}{he} rapid expansion of intelligent technologies across industries and daily life is driven by a growing ecosystem of dedicated \gls{CMOS}-based hardware accelerators offering increasing computing power to support advances in \gls{AI}~\cite{Reuther2022}. However, \gls{CMOS} scaling has
been slowing down for several years, making further enhancements increasingly difficult to achieve~\cite{Radamson2020}. Silicon
photonics has emerged as a promising candidate technology to sustain the growth of computing performance, offering the ability to
leverage the unique properties of light to perform computation. In particular, it has demonstrated superior energy efficiency and
speed compared to traditional \gls{CMOS} circuits when executing core operations required by \gls{AI} workloads such as
\gls{MVM}~\cite{Kitayama2019}. 

Although photonic \gls{AI} accelerators remain in an early stage of development, their potential continues to grow. Ongoing
research steadily advances toward real-world applications, with the introduction of \glspl{PNN} ~\cite{Pavanello2023}, a new class
of \glspl{NN} where key computing primitives, such as matrix multiplications and signal propagation, are performed
totally or partially using light (photons) instead of electricity ~\cite{Tsakyridis2024, Ashtiani2022}. Several works have
successfully demonstrated photonic implementations of various types of \glspl{NN}, including \glspl{FFNN}, \glspl{CNN}, and
\glspl{RNN}, operating fully or partially in the optical domain~\cite{Tsakyridis2024, Khonina2024a}.

The core of many \glspl{PNN} lies in the ability to perform efficient \gls{MVM} operations in the optical domain. To this end, the
\gls{PUM} mesh is a key enabling component. The \gls{PUM} mesh is a photonic circuit with programmable optical elements. The first
\gls{PUM} mesh was an array of \glspl{MZI} devices~\cite{Reck1994} whose collective behavior can be modeled by a unitary transfer
matrix. This mathematical property enables the use of \gls{SVD} to implement general \gls{MVM} optically. In a \gls{PNN}, the
\gls{SVD} allows for the factorization of a weight matrix (e.g., from a fully connected layer) into three components: two unitary matrices
and a diagonal matrix. \gls{PUM} meshes directly map unitary matrices, while optical attenuators or amplifiers typically define the diagonal matrix~\cite{Marchesin2025}. 

\IEEEpubidadjcol
Two primary approaches exist for designing and training a \gls{PNN}. The first involves training a conventional \gls{NN} in
the digital domain and subsequently mapping its weights onto a photonic architecture via \gls{SVD}
~\cite{Tsakyridis2024}. The second approach involves an end-to-end training process, where the \gls{PNN} is trained
directly in the photonic domain. This method models the \gls{PNN} considering the parameters and constraints
the photonic circuits impose. While more complex, this latter method is crucial to accurately account for 
device non-idealities, optical noise, and propagation losses, making it closer to real hardware implementations~\cite{Fu2024}.

In this context, designing and training \glspl{PNN} requires advanced tools to efficiently model the photonic circuits involved in their construction. Simulators such as SIMULIA~\cite{simulia}, RSOFT~\cite{rsoft}, and COMSOL~\cite{comsol} are primarily designed for simulating individual photonic components—like waveguides, \glspl{MZI}, and ring resonators, while Simphony~\cite{Ploeg:2020aa} elevates the simulation for circuit-level analysis. However, these tools are not suitable for end-to-end modeling and training of complete \gls{PNN} architectures. To address this problem, specialized simulation frameworks—such
as Photontorch~\cite{Laporte2019a}, Neurophox~\cite{Pai2019}, and
Neuroptica~\cite{Williamson2020}—have been introduced. These tools were designed for early-stage designs with relatively simple
architectures, and they face scalability issues as \glspl{PNN} increases in complexity~\cite{Marchesin2025, Shekhar2024}. As \gls{PUM} meshes
grow, their transfer matrices become increasingly large and computationally demanding, creating a bottleneck for existing
\gls{SOTA} training tools.

This paper presents a new method to accelerate training in large-scale \gls{PUM} efficiently meshes called \emph{Slicing method}. This technique is implemented in LuxIA\footnote{authors will make the framework open source upon acceptance of the paper}, an open-source Python framework based on PyTorch~\cite{Paszke2019},
specifically designed for scalable, fast, and memory-efficient training of next-generation \glspl{PNN}. LuxIA’s performance is benchmarked against
current tools regarding execution time and resource usage. Experiments demonstrate LuxIA’s effectiveness by training four
\glspl{PNN} architectures—Clements~\cite{Clements2016}, Fldzhyan~\cite{Fldzhyan2020}, Clements Bell, and Fldzhyan
Bell~\cite{Bell2021}—on four standard datasets: Iris~\cite{Fisher1936}, Digits~\cite{Seewald}, MNIST~\cite{Lecun1998}, and
Olivetti Faces~\cite{Samaria1994}.

The paper is organized as follows: Section~\ref{sec:background} provides an overview of photonic accelerators,
modeling approaches and surveys existing \gls{SOTA} tools for \gls{PNN} training, emphasizing their strengths and limitations.
Section~\ref{sec:Methods} presents the proposed methodology and introduces the LuxIA framework. Section~\ref{sec:results}
benchmarks the training efficiency and performance of LuxIA against other tools, and explores diverse use cases by training
multiple \glspl{PNN} models on various datasets. Finally, Section~\ref{sec:conclusion} summarizes the contributions of this work and
outlines future directions.

\section{Background \& Related Works} \label{sec:background} 
Photonic accelerators represent a paradigm shift in computing, leveraging the unique properties of light to execute specific
operations at extremely high speed. These systems operate at frequencies reaching the terahertz range while consuming only
milliwatts of power~\cite{Makarenko2023}. This energy efficiency makes photonic accelerators attractive to high-throughput,
power-constrained \gls{AI} applications. 

\begin{figure}[ht]
    \centering
    \subfigure[Hardware Architecture of a \gls{PNN}.]{
        \centering
        \includegraphics[scale=0.32]{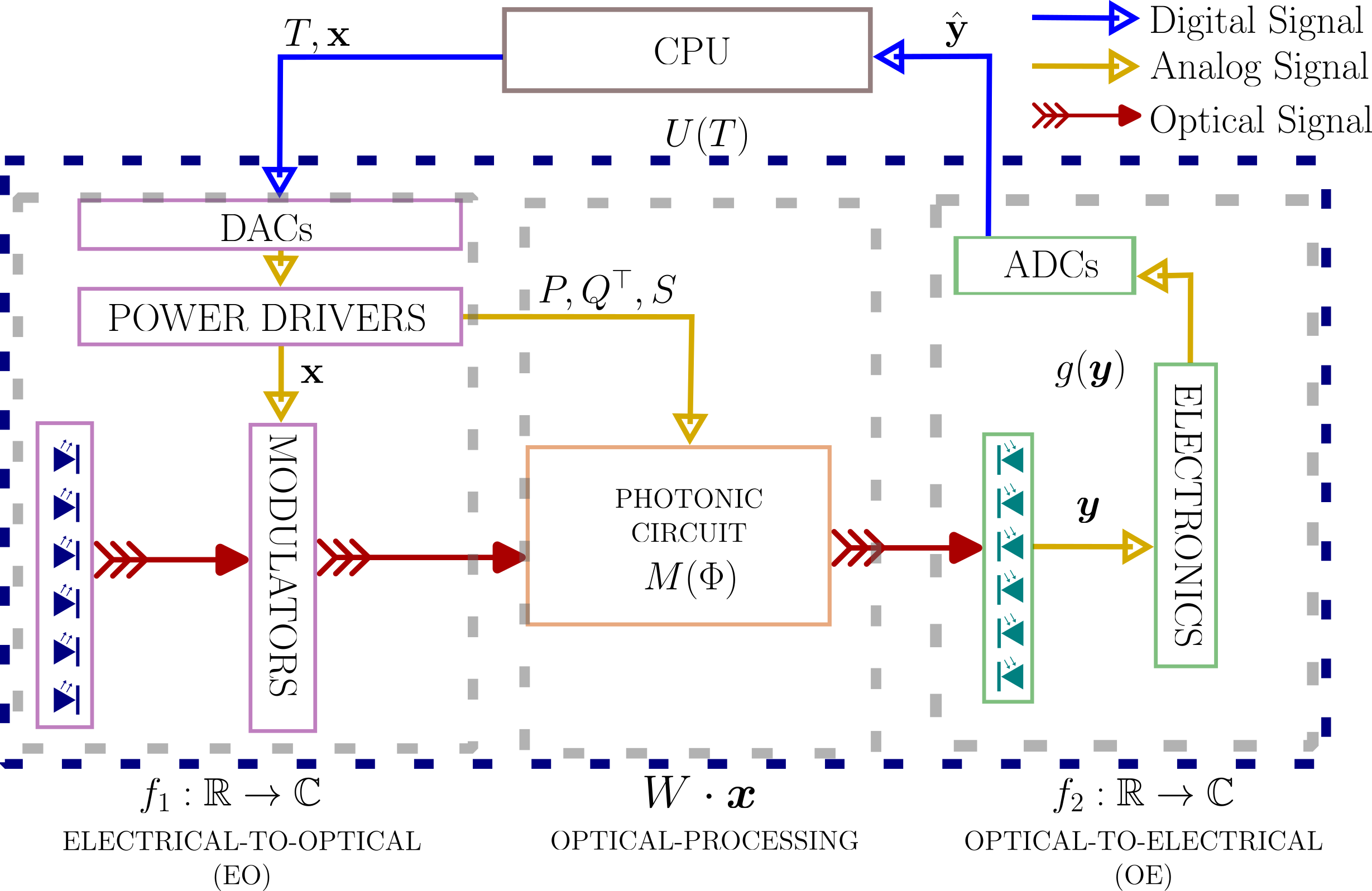}
        \label{fig:hw_architecture}
    }
    \subfigure[Fully Connected Layer of a \gls{NN}.]{
        \centering
        \includegraphics[scale=0.28]{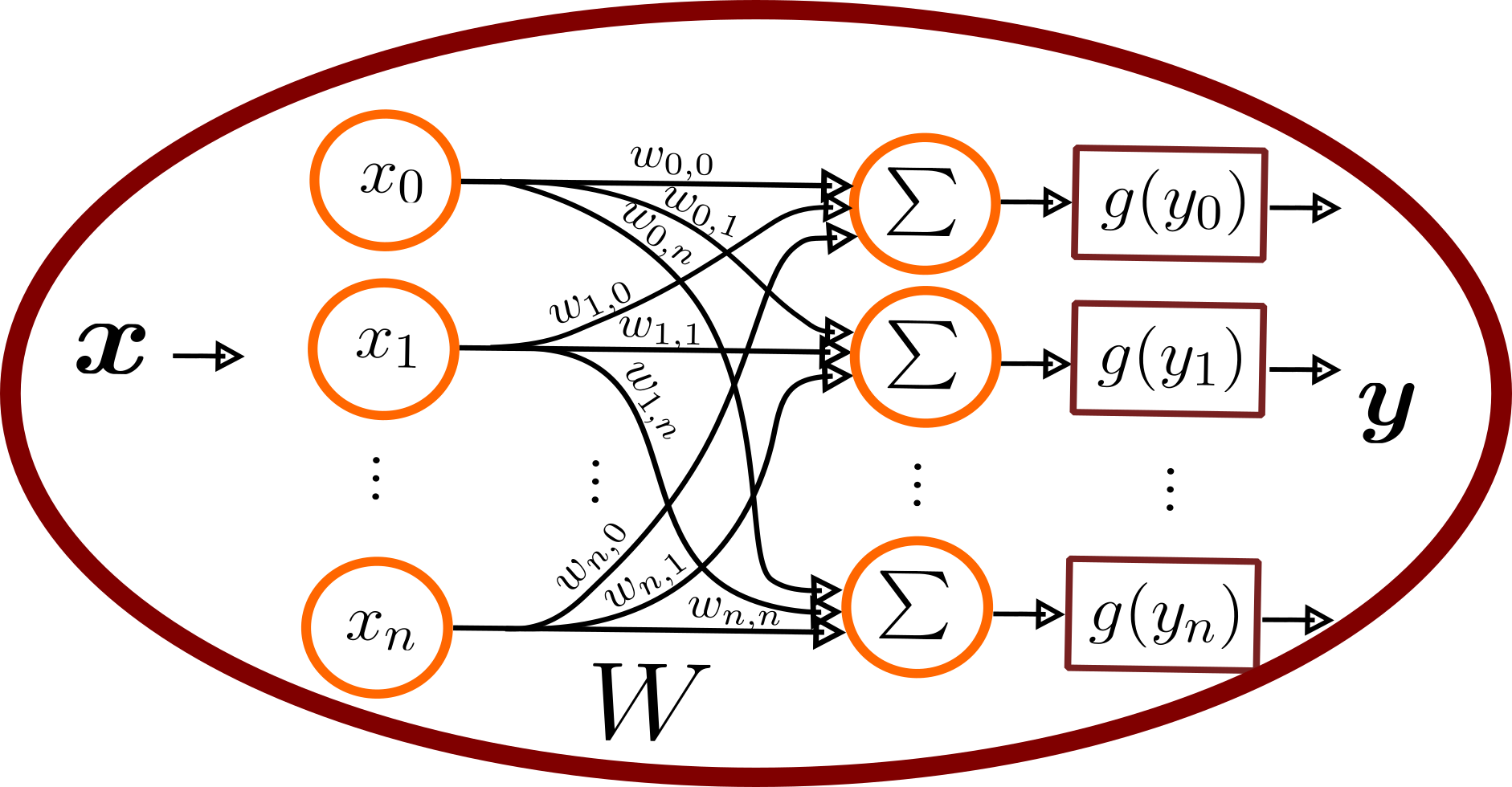}
        \label{fig:nn_architecture}
    }
    \caption{ 
        (a) Hardware implementation on a photonic accelerator and (b) a fully connected layer of a \gls{NN}. In (a), the input
        vector $\mathbf{x}$ is converted to the optical domain for \gls{MVM} computation, and the result is converted back to the
        electronic domain to apply the activation function, producing the output $\mathbf{y}$.
    }
    \label{fig:pnn}
\end{figure}

The general architecture of a photonic accelerator comprises three main processing stages forming a complete pipeline:
\gls{EO}, optical processing, and \gls{OE}~\cite{Andrulis2024}, as illustrated in Figure~\ref{fig:hw_architecture}. The
\gls{EO} provides the interface to convert digital data into optical signals using \glspl{DAC}, precisely controlled lasers,
and modulators. The optical processing stage serves as the computational core, where operations are executed entirely in the
photonic domain, exploiting the properties of light-based computation. Finally, the \gls{OE} stage converts the processed optical
signals back to electrical form using high-speed photodetectors and \glspl{ADC}~\cite{Andrulis2024}.

In \gls{AI} applications, photonic accelerators that perform \gls{NN} operations are called \glspl{PNN}. These systems execute
core \gls{NN} computations optically, offering significant benefits over electronic versions. \glspl{PNN} may be fully optical, with all processing in the optical domain, or opto-electronic, splitting tasks between optical and electronic stages as
needed~\cite{Tsakyridis2024, Khonina2024a}.~\autoref{fig:pnn} illustrates a fully connected layer (Figure
\ref{fig:nn_architecture}) mapped to a photonic accelerator (Figure \ref{fig:hw_architecture}), where the photonic domain handles
the \gls{MVM} operations and the \gls{OE} stage uses electronics for the activation function  $g(y)$, producing the output
$\hat{\mathbf{y}}$ from input $\mathbf{x}$.

\begin{figure}[htb]
    \centering
    \includegraphics[scale=0.4]{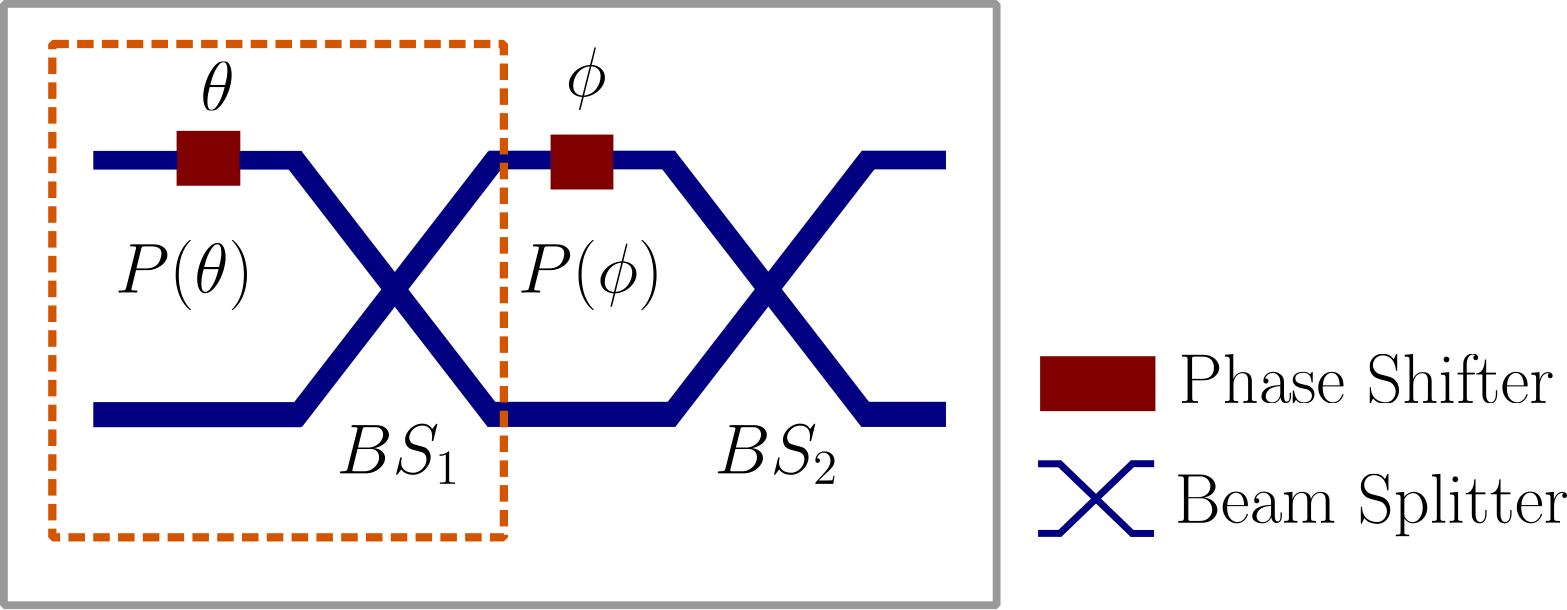}
    \caption{
        A \gls{MZI} is illustrated in the gray rectangle, composed of two beam splitters and two phase shifters ($\theta$ and $\phi$). This
        full \gls{MZI} serves as the single building block in various \gls{PUM} mesh architectures, such as the Clements
        mesh~\cite{Clements2016}. In contrast, the dotted orange rectangle highlights the simpler single building block used in the Fldzhyan mesh~\cite{Fldzhyan2020}, composed of a beam splitter followed by a phase shifter.
    }
    \label{fig:MZI}
\end{figure}

The majority of \glspl{PNN} have a photonic circuit capable of performing \gls{MVM} operations at their core~\cite{Khonina2024a}.
Among various approaches—including \gls{WDM}-based circuits~\cite{Sludds2022} and \gls{PLC}~\cite{Zhou2022}—this work focuses on
\gls{MZI}-based implementations due to their proven scalability and programmability. An \gls{MZI} device (used in the Clements~\cite{Clements2016}), shown in
~\autoref{fig:MZI}, consists of two beam splitters interconnected by waveguides with programmable phase shifters~\cite{Wu2021}.
The beam splitters divide incoming light into two paths, while the phase shifters ($\theta$ and $\phi$ in the Figure) enable programmable
control over optical interference patterns. Taking \autoref{fig:MZI} as an example, the transfer matrix of a single \gls{MZI}
is calculated by composing the matrices of its beam splitters (\autoref{eq:general_bs}) and phase shifters
(\autoref{eq:dig_mat_ps}), resulting in its transfer matrix $M$ as defined in \autoref{eq:mzi_a}.

%
%
%
{\small
\begin{equation} \label{eq:general_bs}
    BS = \frac{1}{\sqrt{2}}
    \begin{bmatrix}
        1 & \mathrm{j} \\
        \mathrm{j} & 1
    \end{bmatrix}
\end{equation}

\begin{equation} \label{eq:dig_mat_ps}
    P(\theta) =
    \begin{bmatrix}
        \mathrm{e}^{\mathrm{j}\theta} & 0 \\
        0 & 1
    \end{bmatrix}
\end{equation}
 \begin{equation} \label{eq:mzi_a}
 \begin{split}
     M & = \underbrace{P(\theta) \cdot BS}_{B_\theta} \cdot \underbrace{P(\phi) \cdot BS}_{B_\phi} \\
      & = \frac{1}{2}
     \begin{bmatrix}
         \mathrm{e}^{\mathrm{j}(\theta+\phi)} - \mathrm{e}^{\mathrm{j}\theta} & \mathrm{j}(\mathrm{e}^{\mathrm{j}(\theta+\phi)} + \mathrm{e}^{\mathrm{j}\theta})\\
         \mathrm{j}(\mathrm{e}^{\mathrm{j}\phi} + 1) & -(\mathrm{e}^{\mathrm{j}\phi} - 1)
     \end{bmatrix}
     \end{split}
 \end{equation}
}

The fact that \glspl{MZI} are programmable makes them suitable for implementing tunable optical
circuits capable of performing diverse operations. \glspl{MZI} can perform \gls{MVM} operations, if they are arranged in
a mesh structure. These mesh structures are known as \gls{PUM} meshes. The term \gls{PUM} refers to the fact that the transfer
matrix of a \gls{PUM} mesh satisfies unitary properties~\cite{Reck1994, Clements2016, Fldzhyan2020, Pai2019a}.

This unitary property enables \glspl{PUM} to perform \gls{MVM} operations by leveraging the mathematical technique known as
\gls{SVD}. The \gls{SVD} factorizes any matrix into three matrices, formally expressed in~\autoref{eq:svd}, where $W$ is the
original matrix, $P$ and $Q$ are unitary matrices, and $S$ is a diagonal matrix containing the singular values. This factorization
is particularly advantageous since any matrix (such as the weight matrix $W$ depicted in Figure~\ref{fig:nn_architecture}) can be
decomposed into two unitary matrices and one diagonal matrix. The unitary matrices $P$ and $Q$ can be directly implemented as the
transfer matrices of the \glspl{PUM} meshes $M$. The diagonal matrix $S$ can be realized using other optical components,
such as attenuators or amplifiers~\cite{Tsakyridis2024,Cheng2021}.

{\small
\begin{equation} \label{eq:svd}
    W = P \cdot S \cdot Q^\top
\end{equation}
}

Reck et al.~\cite{Reck1994} introduced the first \gls{PUM} mesh and proposed a systematic approach to optical \gls{SVD} using
triangular \gls{MZI} arrays. Subsequent developments include the optimized rectangular design by Clements et
al.~\cite{Clements2016} for improved scalability and reduced optical losses, and Fldzhyan's architecture~\cite{Fldzhyan2020} which
utilizes beam splitters and phase shifters instead of full \glspl{MZI}. More recently, Bell et al.~\cite{Bell2021} introduced
further optimizations of these designs. The choice of the \gls{PUM} architecture directly influences how a \gls{PNN} is trained and
its resilience to hardware imperfections.

\gls{PNN} training can be accomplished by two primary methods. The first, post-training mapping, trains a standard \gls{NN} and maps its weights to
photonic hardware via \gls{SVD} into the \gls{PUM} mesh of the \gls{PNN}\cite{Tsakyridis2024}. This simple approach ignores device
non-idealities, causing accuracy loss between simulated and real \gls{PNN}\cite{Banerjee2021}. The second, end-to-end photonic
training, optimizes the tunable parameters in the \gls{PNN}, such as programmable phase shifters in the \gls{PUM} meshes.
Though more complex, the latter models non-idealities like optical noise, losses, and manufacturing variations, yielding training that
better matches hardware performance~\cite{Fu2024}.

Effective end-to-end training requires accurate modeling of the \gls{PNN}. Such training is typically achieved using the transfer matrix
method, which provides the mathematical relationship between the system's inputs and outputs~\cite{Mackay2020}. The transfer
matrix method is preferred for \gls{PNN} modeling due to its compact representation, compatibility with optimization algorithms,
and computational efficiency~\cite{Laporte2019,Laporte2019a}. Starting from this model, the end-to-end training process follows the
conventional four stages required to train a \gls{NN}:
forward pass (Equation~\ref{eq:forward_pass}), loss computation (Equation~\ref{eq:loss_function}), and backward propagation
(Equations~\ref{eq:backward_pass_0}–\ref{eq:backward_pass_2}).

{\small
\begin{equation} \label{eq:forward_pass}
    \hat{\mathbf{y}} = U(T)\cdot\mathbf{x}
\end{equation}

\begin{equation} \label{eq:loss_function}
    \mathscr{L} = f\bigl(\mathbf{y},\hat{\mathbf{y}}\bigr)
\end{equation}

\begin{equation} \label{eq:backward_pass_0}
    \delta = \nabla_{\mathbf{\hat{y}}} \mathscr{L}
\end{equation}

\begin{equation} \label{eq:backward_pass_1}
    \delta^{[l]} = (T^{[l+1]})^\top \delta^{[l+1]}
\end{equation}

\begin{equation} \label{eq:backward_pass_2}
    T^{[l]} = T^{[l]} - \eta \cdot \delta^{[l]} (\hat{\mathbf{y}}^{[l-1]})^\top
\end{equation}
}

Here, the input $\mathbf{x}$ is transformed by the system's end-to-end transfer matrix $U(T)$ to produce the output
$\hat{\mathbf{y}}$. The loss $\mathscr{L}$ is computed between the prediction and the true label $\mathbf{y}$. The gradient of the loss
$\delta$ is then propagated backward through the system to update the trainable parameters $T$. As shown in Equations
\ref{eq:backward_pass_1} and \ref{eq:backward_pass_2}, this is a layer-wise process where the superscript index $l$ refers
to a specific "subsystem" in the end-to-end model\footnote{The term "layer" in the context of backpropagation (indexed by $l$)
should be understood as a computational stage or subsystem within the \gls{PNN}, distinct from a conventional neural
network layer, e.g., a fully connected layer}. The parameters $T^{[l]}$ of each subsystem are updated using the related gradient
$\delta^{[l]}$ and the learning rate $\eta$.

\begin{figure}[ht]
    \centering
    \subfigure[Layers configuration]{\includegraphics[scale=0.65]{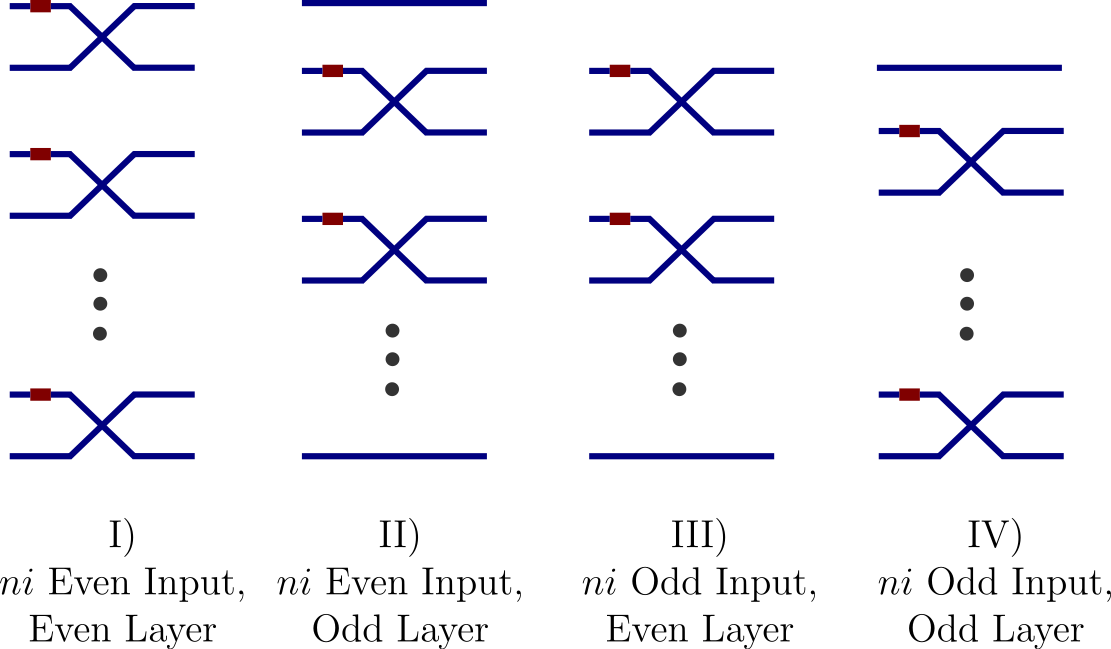} \label{fig:Fldzhyan_lay}}
    \subfigure[Photonic mesh]{\includegraphics[scale=0.55]{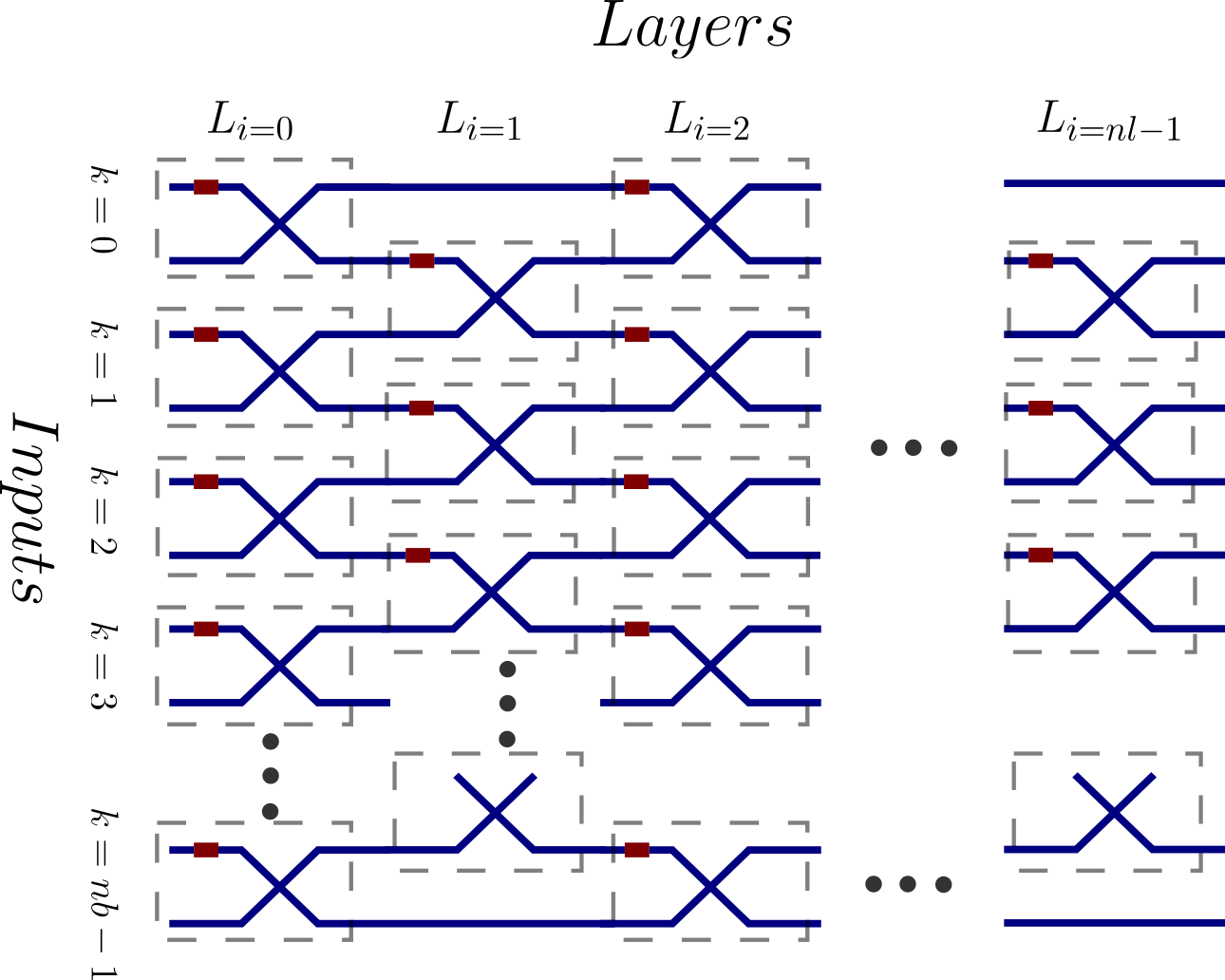} \label{fig:Fldzhyan_arch}}
    \caption{Fldzhyan mesh structure showing (a) single block, (b) mesh layers, and (c) architecture.}
    \label{fig:Fldzhyan}
\end{figure}

The system matrix $U(T)$ is constructed by multiplying the transfer matrices of its constituent subsystems. These can include
input modulators, output detectors, \gls{PUM} meshes, etc. It is important to note that each subsystem can have its  
trainable parameters, as illustrated in the case of \gls{PUM} meshes, where the transfer matrix $M$ comprises
the trainable phase shift parameters. As seen before, while the calculation for a single component is straightforward, the transfer matrix of a full system is far more
complex, especially for large \gls{PUM} meshes. Harris et al.~\cite{Harris2017} were able to build a photonic accelerator with 88
\glspl{MZI}, showcasing the feasibility of large-scale integration, while modern \gls{PNN} architectures sytematically require the use of larger
\gls{PUM} meshes with new optical components~\cite{Tsakyridis2024, Khonina2024a, Marchesin2025, Shekhar2024}. Consequently,
training \glspl{PNN} with large \gls{PUM} meshes presents a significant scalability challenge due to the computational cost of
evaluating their transfer matrices, leading to long training times~\cite{Destras2023}. To illustrate this complexity, consider a
photonic mesh structure, such as the Fldzhyan mesh illustrated in \autoref{fig:Fldzhyan}.
The analysis begins by identifying the fundamental repeating unit, i.e., the single block, and computing its transfer matrix,
denoted by $B$. Next, the structure is decomposed into different mesh layers, and the transfer matrix for each mesh layer is computed. The final step is to compute the transfer matrix of the entire mesh, denoted by $M$, by
multiplying the transfer matrices of all mesh layers. Key parameters such as the number of inputs ($ni$), the number of layers ($nl$), and the number of single blocks per layer ($nb$) must be identified. These parameters are essential throughout all steps.

The transfer matrix of a single block $B_{i,k}$---uniquely identified by the two-level subindex $(i,k)$ where $i$ corresponds to the mesh layer number and $k$
indicates the block number within that layer---is obtained by multiplying the transfer matrices of its internal
components. As illustrated in gray dashed rectangles of ~\autoref{fig:Fldzhyan}, single blocks in the Fldzhyan mesh comprise a beam splitter and a phase
shifter. 
Hence, the transfer matrix of the block $B_{i,k}$ can be calculated as the product of the transfer
matrix of the beam splitter, $BS$, and the phase shifter, $P(\theta_{i,k})$ where $\theta_{i,k}$ denotes the phase shift of block
$B_{i,k}$. Then, the transfer matrix $B_{i,k}$ is computed as shown
in~\autoref{eq:fldzhyan_sb}, with indices $i \in [0, nl-1]$ and $k \in [0, nb-1]$.

{\small
\begin{equation}     
        B_{i,k} = \underbrace{\frac{1}{\sqrt{2}}
                \begin{bmatrix}
                    \mathrm{e}^{\mathrm{j}\theta_{i,k}} & \mathrm{j} \\
                    \mathrm{j}\mathrm{e}^{\mathrm{j}\theta_{i,k}} & 1 \\
                \end{bmatrix}}_{BS~\cdot~P(\theta_{i,k})} 
                \label{eq:fldzhyan_sb}
\end{equation}
}

Once the transfer matrix of all individual blocks has been computed, the next step is to calculate the mesh layer transfer
matrices, denoted by $L$. First, identify all single blocks that form a given layer, with their transfer matrices denoted as
$B_{i,k}$ for all $k$. Since each layer is a system in which the number
of outputs equals the number of inputs, the mesh layer transfer matrix $L$ must be square, with dimensions $ni \times ni$.

Each mesh layer transfer matrix $L_i$ is then constructed by placing the transfer matrices of its constituent
blocks $B_{i,k}$ along the diagonal, while setting all off-diagonal elements to zero, as shown in
~\autoref{eq:mesh_layer}. 

{\small
\begin{equation} 
    L_{i} \underset{i \in [0, nl-1]}= 
    \begin{bmatrix}
        B_{i,0} & 0 & \cdots & 0 \\
        0 & B_{i,1} & \cdots & 0 \\
        \vdots & \vdots & \ddots & \vdots \\
        0 & 0 & \cdots & B_{i,nb-1}
    \end{bmatrix} \label{eq:mesh_layer}
\end{equation}
}

Regarding the Fldzhyan mesh structure, the mesh layers are arranged in four different configurations as shown in Figure \ref{fig:Fldzhyan_lay}, depending on two parameters: the number of inputs $ni$ and the index of the current layer $i$.
Equations \ref{eq:fldzhyan_layer_even} and \ref{eq:fldzhyan_layer_odd} show the development of the transfer matrix when $ni$ is even. 

{\small
\begin{equation}\label{eq:fldzhyan_layer_even}
    L_{i} \underset{ i \in 2\mathbb{Z}}{=} \frac{1}{\sqrt{2}}
    \underbrace{\begin{bmatrix}
        \mathrm{e}^{\mathrm{j}\theta_{i,0}}  & \mathrm{j} & \dots  & 0     & 0 \\
        \mathrm{j}\mathrm{e}^{\mathrm{j}\theta_{i,0}} & 1 & \ddots & \vdots & \vdots \\
        \vdots  &\vdots &\ddots  & \mathrm{e}^{\mathrm{j}\theta_{i,nb-1}} & \mathrm{j} \\
        0       &   0   & \dots  & \mathrm{j} \mathrm{e}^{\mathrm{j}\theta_{i,nb-1}} & 1
    \end{bmatrix}}_{\mathbb{L}}
    \end{equation}
}


{\small
\begin{equation}\label{eq:fldzhyan_layer_odd}
L_{i} \underset{i \in (2\mathbb{Z}+1)}{=}\frac{1}{\sqrt{2}}     
    \begin{bmatrix}
        1 & 0 &  \dots  & 0 & 0\\
        \vdots &   & \mathbb{L}  &  & \vdots \\
        0 & 0 &  \dots  & 0 & 1\\
    \end{bmatrix}
\end{equation}
}

When the number of inputs $ni$ is odd, a similar procedure is followed, and the final Equations are shown in 
Equations \ref{eq:fldzhyan_olayer_even} and \ref{eq:fldzhyan_olayer_odd}.

{\small
\begin{equation}\label{eq:fldzhyan_olayer_even}
L_{i} \underset{i \in 2\mathbb{Z}}{=}\frac{1}{\sqrt{2}}     
    \begin{bmatrix}
         \mathbb{L} & &    0\\
                    & &    \vdots\\
         0          & \dots      &     1\\
    \end{bmatrix}
\end{equation}

\begin{equation}\label{eq:fldzhyan_olayer_odd}
L_{i} \underset{i \in (2\mathbb{Z}+1)}{=}\frac{1}{\sqrt{2}}     
    \begin{bmatrix}
         1      & \dots & 0\\
        \vdots& &       &  \\
        0       &       &\mathbb{L}\\
    \end{bmatrix}
\end{equation}
}

The final step is to compute the complete photonic mesh structure's transfer matrix $M$ by multiplying the mesh layers' transfer matrices, as given by~\autoref{eq:fldzhyan_arch}.

{\small
\begin{equation} 
    M = L_{0} \cdot L_{1} \cdot L_{2} \cdot \ldots \cdot L_{nl-1} \label{eq:fldzhyan_arch}
\end{equation}
}
The transfer matrix method provides a rigorous and widely used framework for modeling and training \glspl{PNN}. However, as shown in \autoref{eq:fldzhyan_sb}, the number of matrix multiplications required to construct the single block transfer matrices increases with the number of photonic devices of the circuit. Additionally, \autoref{eq:fldzhyan_layer_even} highlights that for \glspl{PNN} with large \gls{PUM} meshes, the dimensions of each mesh layer transfer matrices scale quadratically with the number of inputs. Finally, \autoref{eq:fldzhyan_arch} shows that the transfer matrix of the entire mesh is the product of all mesh layer transfer matrices. This introduces the following key limitations when training \glspl{PNN}:

\begin{IEEEitemize}
    \item \emph{Computational Overhead:} As input dimensionality grows, the matrix size and number of required multiplications
        increase quadratically. Furthermore, the sparse structure of these matrices leads to useless computations, as many
        operations involve multiplication by off-diagonal zero elements.

    \item \emph{High Memory Usage:} Storing large, dense transfer matrices imposes a substantial memory burden. This is
        particularly problematic when training \glspl{PNN} on \glspl{GPU}, where available memory is often a critical constraint.

    \item \emph{Gradient Bottleneck:} This critical issue arises during the weights update process of the backpropagation
        algorithm. As described in \autoref{eq:backward_pass_2} and the preceding gradient computations in Equations
        \ref{eq:backward_pass_0} and \ref{eq:backward_pass_1}, updating the parameters of the full system transfer matrix
        $U(T)$ requires evaluating the gradient of the loss $\mathscr{L}$ with respect to all trainable parameters $T$. 

        The key challenge is that the mesh transfer matrix $M$, which is part of $U(T)$, represents a highly entangled many-to-one mapping: the effect of each phase shift is inseparably combined with contributions from all other parameters. Because of this entanglement, the loss gradient concerning any single parameter cannot be isolated from the final output alone.  

        Consequently, each backward pass necessitates reconstructing the full mesh transfer matrix $M$ from its constituent layer matrices $L_i$ to correctly compute gradients for all parameters. This repeated and costly reconstruction creates a
        significant computational bottleneck, especially for large-scale meshes.
\end{IEEEitemize}

Despite growing interest in \glspl{PNN}, the availability of frameworks for their end-to-end training remains limited. Among the
most notable Python-based tools are Photontorch~\cite{Laporte2019a}, Neurophox~\cite{Pai2019}, and Neuroptica~\cite{Williamson2020}. These frameworks differ in backend technologies, hardware acceleration
capabilities, and development activity. Table~\ref{tab:related_tools} summarizes their technical characteristics.

\begin{table}[ht]
    \centering
    \caption{Technical summary of \gls{PNN} simulation tools}
    \label{tab:related_tools}
    \begin{tabular}{| c | c | c | c |}
        \hline
        \textbf{Tool} & \textbf{Backend} & \textbf{GPU Support} & \textbf{Github Last Commit} \\
        \hline
        Photontorch & PyTorch & Yes & 2022 \\
        \hline
        Neurophox   & TensorFlow & Yes & 2021 \\
        \hline
        Neuroptica &  NumPy & No & 2020 \\
        \hline
    \end{tabular}
\end{table}

Photontorch is a simulation framework built on PyTorch~\cite{Paszke2019}, offering automatic differentiation and
\gls{GPU} acceleration to support gradient-based training of \glspl{PNN}. It has been used to simulate architectures such as
\glspl{CROW}~\cite{Poon2004} and \glspl{PNN} with \glspl{PUM} \glspl{MZI} for tasks like MNIST classification.
Photontorch includes prebuilt \gls{PUM} configurations, such as the Clements mesh~\cite{Clements2016}, and provides a
user-friendly interface for building custom photonic circuits. However, training large-scale circuits remains time-consuming,
often requiring hours to optimize even a few hundred parameters~\cite{Destras2023}.

Neurophox, developed by Pai et al., is a TensorFlow-based framework focused on rectangular and triangular \gls{PUM}
meshes composed of \glspl{MZI} and phase shifters. It uses the transfer matrix method to model photonic components and supports
\gls{GPU}-accelerated backpropagation. While the framework benefits from integration with TensorFlow's optimizers and includes
mesh-specific enhancements~\cite{Williamson2020a}, its training performance degrades as the network depth and mesh size
increase~\cite{Destras2023}.

Neuroptica is a NumPy-based framework that implements a custom backpropagation algorithm for PNNs, based on the adjoint method described in ~\cite{Hughes2018}. Like
the others, it relies on the transfer matrix method to simulate beam splitters, \glspl{MZI}, and phase shifters. Neuroptica allows users to build networks at different levels of abstraction. However, it lacks \gls{GPU} support and exhibits particularly slow training times as model complexity grows~\cite{Destras2023}.

Across all three tools, the most significant drawback is their slow training speed, which becomes a major bottleneck in practical applications. As emphasized in Destras et al.~\cite{Destras2023}, the simulation and optimization of even moderate-scale \gls{PNN} architectures can take hours, largely due to the computational overhead introduced by differentiating through photonic circuits modeled with the transfer matrix method. While the frameworks vary in abstraction and backend support, none offer efficient alternatives to mitigate this core issue. Although a lack of recent updates or long-term maintenance can affect
compatibility, these concerns are often manageable through containerization solutions such as Conda, Docker, or
Singularity~\cite{Wenhao2020}. In contrast, the inefficiency in training remains a fundamental limitation — and a primary obstacle to the scalable adoption of \gls{PNN} simulation frameworks.

\section{Methodology} \label{sec:Methods}
To address the significant computational challenges posed by training large-scale photonic mesh
networks, LuxIA implements the Slicing Method, a novel algorithm conceptually inspired by the physical organization of photonics accelerators where the light travels sequentially
through one optical element at a time. The Slicing Method decomposes the complete photonic mesh into a
sequence of computational "windows." Unlike the conventional transfer matrix method, which requires assembling and storing a
single large matrix representing the entire system, the Slicing Method computes the signal propagation incrementally. This is
achieved through localized matrix-vector operations, substantially reducing both memory requirements
and computational complexity.

\begin{figure}[htb]
    \centering
    \includegraphics[scale=0.48]{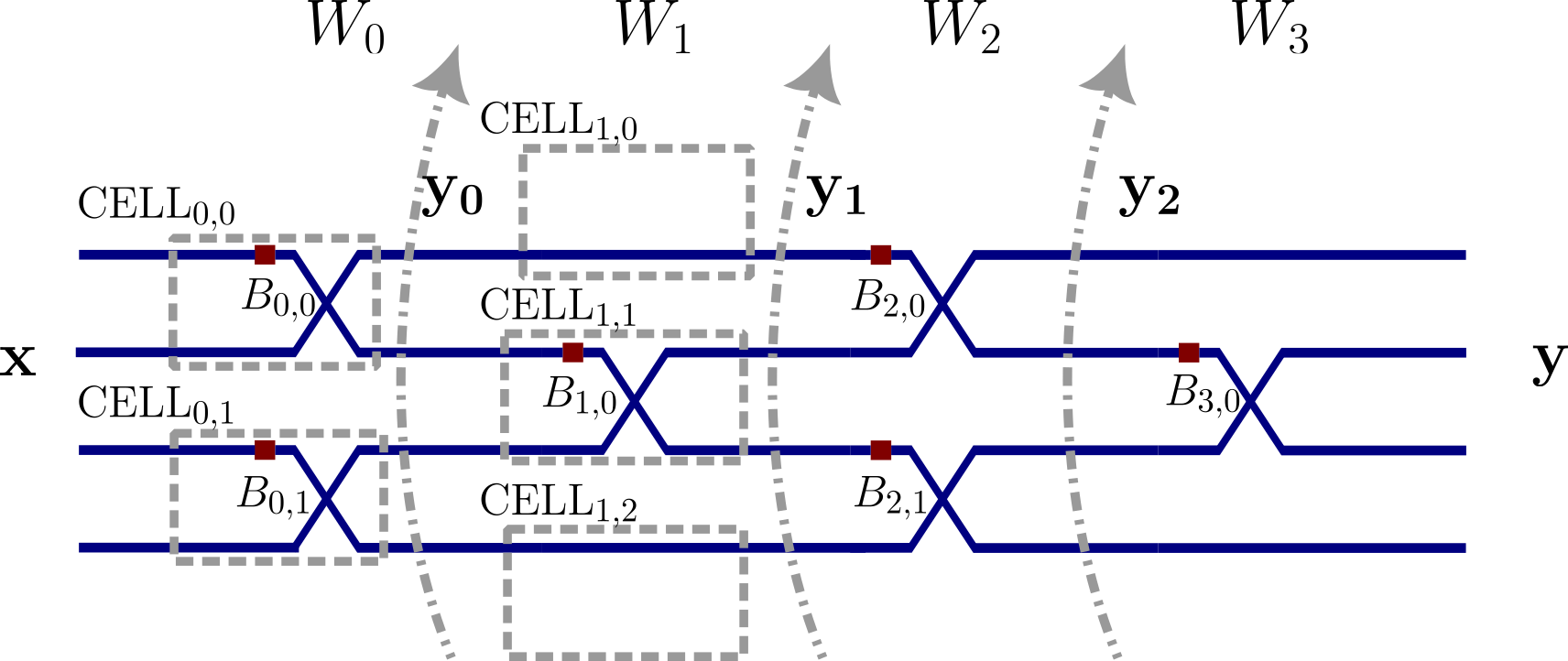}
    \caption{Decomposition of the Fldzhyan mesh structure into sequential processing windows. Physical blocks ($B_{i,k}$) are
    mapped to computational cells ($\text{CELL}_{w,c}$) for processing.}
    \label{fig:fldzhyan_mesh_slicing_decomposition}
\end{figure}

The methodology is formalized through the following systematic steps:

\begin{enumerate} 
    \item \emph{Window Partitioning:} The mesh is computationally divided into $nw$ processing windows, where for many common
        architectures $nw$ equals the number of physical layers, $nl$. Each window, denoted $W_w$, contains a group of optical
        operations that can be computed independently within that slice.

    \item \emph{Cell Identification:} This step introduces the core abstraction of the Slicing Method: the \emph{cell}. The
        purpose of the cell is to provide a uniform computational interface for all possible elements within a window. Therefore,
        every element—a single block or a simple bypass connection—is encapsulated within a cell object. 
        The single block is exactly identified by its layer $i \in [0, nl-1] \cap \mathbb{Z}$ and block number $k \in [0, nb-1] \cap
        \mathbb{Z}$ as $B_{i,k}$, and mapped to an active cell. Crucially, a path that does not contain a block is mapped to a
        bypass cell. Each cell, denoted as $\text{CELL}_{w,c}$, stores an internal flag, \texttt{is\_active}, that allows the
        processing algorithm to treat all elements uniformly:
    \begin{itemize}
        \item An \emph{active cell} (\texttt{is\_active = true}) encapsulates a $2 \times 2$ optical block and contains its
            transfer matrix (e.g., $B_{i,k}$).
        \item A \emph{bypass cell} (\texttt{is\_active = false}) encapsulates a single pass-through connection and requires no
            further information.
    \end{itemize}
    This encapsulation is key to the framework's generality. As illustrated in \autoref{fig:fldzhyan_mesh_slicing_decomposition},
    window $W_0$ maps the single blocks $B_{0,0}$ and $B_{0,1}$ to two active cells ($\text{CELL}_{0,0}$, $\text{CELL}_{0,1}$),
    while window $W_1$ is represented as an ordered list of three cells: a bypass cell ($\text{CELL}_{1,0}$), an active cell
    containing $B_{1,0}$ ($\text{CELL}_{1,1}$), and another bypass cell ($\text{CELL}_{1,2}$).

    \item \emph{Sequential Signal Propagation:} An input vector $\mathbf{x} \in \mathbb{C}^{ni}$ is fed into the first window,
        $W_0$. The output vector from window $W_w$, denoted $\mathbf{y}_w$, becomes the input vector for the subsequent window,
        $W_{w+1}$, creating a computational cascade that mirrors the physical flow of light.
\end{enumerate}

This section formalizes the slicing concept into the generalized computational framework detailed in Algorithm
\ref{alg:photonic_mesh_processing}. The power of this framework lies in its uniform treatment of disparate physical components,
made possible by the cell abstraction. Therefore, the core `process\_window` function is agnostic to the specific mesh topology
(e.g., Fldzhyan vs. Clements); it only needs to process a simple, ordered list of cell objects.

The high-level `process\_mesh` function orchestrates the overall signal flow. It initializes the computation with the mesh input
vector and then iterates through each window, sequentially updating the signal vector by calling `process\_window` for each one.

The core logic resides within the `process\_window` function. This function accepts an input vector and the window's list of cells.
A single port index, $p$, tracks the current position in the input and output vectors. The function iterates through the cells,
and its logic branches based on the \texttt{is\_active} flag of each cell: 
\begin{itemize} 
    \item If the cell is active, it consumes two inputs, performs a $2 \times 2$ matrix-vector multiplication using its stored
        transfer matrix, and produces two outputs. The port index $p$ is then advanced by two. 
    \item If the cell is a bypass, it consumes one input, performs an identity copy, and produces one output. The port index $k$
        is advanced by one. 
\end{itemize} 
For an active cell representing the single block $B_{i,k}$, the mathematical operation is a transformation from $\mathbb{C}^2
\to \mathbb{C}^2$, as depicted in \autoref{eq:active_block_processing}.

{\small
\begin{equation} 
\label{eq:active_block_processing}
\begin{bmatrix} y_{p} \\ y_{p+1} \end{bmatrix} = B_{i,k} \begin{bmatrix} x_{p} \\ x_{p+1} \end{bmatrix}
\end{equation}
}

\autoref{eq:bypass_processing}, shows the identity transformation from $\mathbb{C} \to \mathbb{C}$, for a bypass cell. 
{\small
\begin{equation} \label{eq:bypass_processing}
y_p = x_p
\end{equation}
}

\begin{algorithm}[tb]
\caption{Photonic Mesh Processing (Signal Propagation)}
\label{alg:photonic_mesh_processing}
\begin{algorithmic}[1]
{\scriptsize
\Function{\text{process\_mesh}}{$\mathbf{x}$, \text{mesh}}
    \State $\mathbf{y} \leftarrow \mathbf{x}$
    \For{\textbf{each} \text{window} \textbf{in} \text{mesh.windows}}
        \State $\mathbf{y} \leftarrow \Call{\text{process\_window}}{\mathbf{y}, \text{window}}$
    \EndFor
    \State \Return $\mathbf{y}$
\EndFunction

\vspace{1em}

\Function{\text{process\_window}}{$\mathbf{x}$, \text{window}}
\State $\mathbf{y} \leftarrow \mathbf{x}$ \Comment{This automatically do the bypass cells}
    \State $p \leftarrow 0$
    
    \For{\textbf{each} \text{cell} \textbf{in} \text{window.cells}}
    \If{\text{cell.is\_active}}
        \State $y_0, y_1 \leftarrow \text{cell.TransferMatrix} \cdot [\mathbf{x}[p],\mathbf{x}[p+1]]^\top$
        \State $\mathbf{y}[p] \leftarrow y_0$
        \State $\mathbf{y}[p+1] \leftarrow y_1$
        \State $p\leftarrow p + 2$
    \Else
        \State $p \leftarrow p + 1$ \Comment{Bypass cell, identity copy}
    \EndIf
    \EndFor
    
    \State \Return $\mathbf{y}$
\EndFunction
}
\end{algorithmic}
\end{algorithm}


To make this generalized framework concrete, it is applied to the 4$\times$4 Fldzhyan mesh from
\autoref{fig:fldzhyan_mesh_slicing_decomposition}, which has $ni = 4$ inputs and $nl = 4$ layers. The mapping of single blocks to
computational cells for its four windows ($nw=4$) is concisely summarized in \autoref{tab:fldzhyan_cell_configuration}.

\begin{table}[ht]
\centering
\caption{Cell configuration for the Fldzhyan 4$\times$4 mesh windows. The cell index $c$ runs sequentially in each window.}
\label{tab:fldzhyan_cell_configuration}
\begin{tabular}{|c|c|c|c|}
\hline
\textbf{Window ($w$)} & \textbf{Cell 0 ($c=0$)} & \textbf{Cell 1 ($c=1$)} & \textbf{Cell 2 ($c=2$)} \\
\hline
$W_0$ & Active ($B_{0,0}$) & Active ($B_{0,1}$) & - \\
\hline
$W_1$ & Bypass & Active ($B_{1,0}$) & Bypass \\
\hline
$W_2$ & Active ($B_{2,0}$) & Active ($B_{2,1}$) & - \\
\hline
$W_3$ & Bypass & Active ($B_{3,0}$) & Bypass \\
\hline
\end{tabular}
\end{table}

The fundamental active element in the Fldzhyan architecture is a phase-shifted beam splitter. For each active cell $\text{CELL}{w,c}$, which corresponds to a single processing block $B_{i,k}$, the generic matrix operation described in \autoref{eq:active_block_processing} simplifies to a specific form governed by a single phase shift parameter, $\theta_{i,k}$.

The inputs to block $B_{i,k}$, denoted $x_p$ and $x_{p+1}$, are taken from the outputs of the previous window $W_{w-1}$—specifically, $y_{w-1,p}$ and $y_{w-1,p+1}$. The outputs of the current block, $y_p$ and $y_{p+1}$, become the outputs of the current window $W_w$, labeled $y_{w,p}$ and $y_{w,p+1}$.

The transformation performed by the active cell is defined by the transfer matrix shown in \autoref{eq:fldzhyan_coupler_processing}:

{\small
\begin{equation} \label{eq:fldzhyan_coupler_processing}
\begin{aligned}
y_{w,p} &= \frac{1}{\sqrt{2}} \left( e^{\mathrm{j}\theta_{i,k}} y_{w-1,p} + \mathrm{j} y_{w-1,p+1} \right) \\
y_{w,p+1} &= \frac{1}{\sqrt{2}} \left( \mathrm{j} e^{\mathrm{j}\theta_{i,k}} y_{w-1,p} + y_{w-1,p+1} \right)
\end{aligned}
\end{equation}
}

This equation governs the behavior of any active cell in the architecture. If a cell is inactive (i.e., \texttt{cell.is\_active} is false), a simple bypass transformation is applied instead, as defined in \autoref{eq:bypass_processing}, where the input $x_p = y_{w-1,p}$ is passed directly to the output $y_p = y_{w,p}$.


A crucial advantage of the Slicing Method is its inherent compatibility with gradient-based optimization algorithms, the backbone
of modern \gls{ML}. The localized and sequential nature of the computation enables an efficient implementation of
backpropagation without needing to assemble the full, system-wide transfer matrix.

The gradient of a loss function $\mathscr{L}$ concerning a tunable parameter $\theta_{i,k}$ is calculated via the chain rule.
This process propagates the error signal backward through the mesh, window by window. The update for a specific
parameter depends only on local information:

{\small
\begin{equation} \label{eq:slicing_gradient_chain}
    \frac{\partial \mathscr{L}}{\partial \theta_{i,k}} = \frac{\partial \mathscr{L}}{\partial \mathbf{y}_{w}} \cdot \frac{\partial \mathbf{y}_w}{\partial \theta_{i,k}}
\end{equation}
}
In this expression, $\frac{\partial \mathscr{L}}{\partial \mathbf{y}_{w}}$ is the gradient of the loss concerning the output of the window $w$ containing the parameter propagated backward from the final layer. The term $\frac{\partial \mathbf{y}_w}{\partial
\theta_{i,k}}$ is the local gradient, which is non-zero only for the outputs of the specific cell, $\text{CELL}_{w,c}$, that
contains the parameter $\theta_{i,k}$.

Using the Fldzhyan block model from \autoref{eq:fldzhyan_coupler_processing}, 
the local partial derivatives concerning its
phase parameter $\theta_{i,k}$ are:

{\small
\begin{equation} \label{eq:local_phase_gradient}
    \frac{\partial y_{w,p}}{\partial \theta_{i,k}} = \frac{\mathrm{j}}{\sqrt{2}} \mathrm{e}^{\mathrm{j}\theta_{i,k}} y_{w-1,p}
\end{equation}
\begin{equation} \label{eq:local_coupling_gradient}
    \frac{\partial y_{w,p+1}}{\partial \theta_{i,k}} = -\frac{1}{\sqrt{2}} \mathrm{e}^{\mathrm{j}\theta_{i,k}} y_{w-1,p}
\end{equation}
}
This locality is the key to the method's computational efficiency. The gradient calculation only requires the input to the cell
($y_{w-1,p}$) which was stored during the forward pass, avoiding any need for large matrix manipulations. Consequently, the
computational complexity for both the forward pass (inference) and the backward pass (gradient calculation) scales linearly with
the number of tunable parameters, which is proportional to $ni \cdot nl$ for a typical mesh and not quadratically to the number of
inputs and layers, as in the transfer matrix method.

\section{Experimental Design}\label{sec:results}
This section evaluates LuxIA by comparing it with other \gls{SOTA} tools in terms of training performance and hardware resource
utilization. Subsection~\ref{subsec:results_frameworks} assesses LuxIA's functionality by verifying that it produces training and validation curves comparable to existing tools when applied to identical \glspl{PNN} and datasets.
Subsection~\ref{subsec:hardware_comparison} investigates hardware efficiency regarding memory usage and execution time, and
Subsection~\ref{subsec:use_cases} demonstrates LuxIA's robustness by training various \glspl{PNN} with different
\glspl{PUM} meshes and datasets.

\begin{figure}[ht]
    \centering
    \includegraphics[scale=0.55]{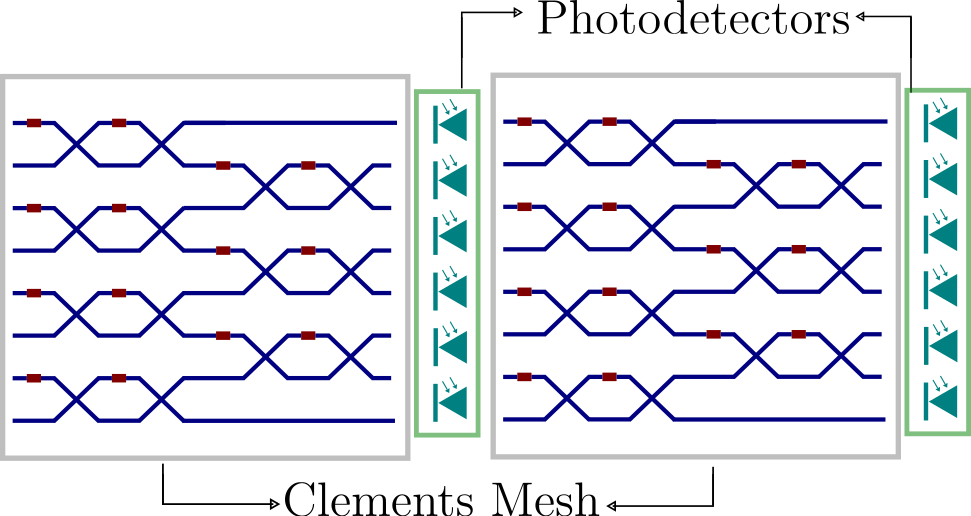}
    \caption{
        \gls{PNN} used for the benchmark experiments. The circuit consists of two sets of Clements meshes followed by a photodetector layer.
        The mesh size varies depending on the experiment: for Subsection~\ref{subsec:results_frameworks}, the size is fixed to
        $N=64$; for Subsection~\ref{subsec:hardware_comparison}, the size depends on the scenario. For the Batch Size-Dependent
        scenario, the size is fixed to $N=800$, while for the \textit{N}-Dependent experiment, $N$ ranges from 100 to 900 in steps
        of 200. 
    }
    \label{fig:benchmark_circuit}
\end{figure}

The experiments in subsections ~\ref{subsec:results_frameworks} and \ref{subsec:hardware_comparison} use 
the \gls{PNN} shown in Figure~\ref{fig:benchmark_circuit}, and the reference dataset used is the
Digits~\cite{Seewald}. This dataset contains 1,797 grayscale images of handwritten digits (0--9), each sized $8 \times 8$. We split the dataset into training (70\%), validation (10\%), and testing (20\%) subsets for all experiments. We selected this dataset because its similarity to the MNIST dataset~\cite{Lecun1998} makes it suitable for \gls{PNN} research, while also keeping the dataset size contained.

To serve as a hardware reference, all experiments run on a workstation equipped with an AMD Ryzen 9 7950X processor featuring 16 cores and 32 threads, alongside 64 GB RAM and an NVIDIA RTX A4000 \gls{GPU} with 16 GB VRAM.

\subsection{Comparison with Other Frameworks as a Training Tool} \label{subsec:results_frameworks}
Validating LuxIA's functionality ensures its effectiveness in training \glspl{PNN}, and it is
performed by comparing LuxIA against other \gls{SOTA} tools. The comparison focuses on training dynamics, rather
than merely comparing outputs. This approach is crucial because the training process 
involves multiple iterations, optimizer behavior, and gradient computations, which are not captured by static input-output
comparisons. For this reason, the evaluation is conducted over a complete training pipeline, including training, validation, and testing phases. 

\begin{figure*}[ht]
    \centering
    \subfigure[Training loss]{
        \centering
        \includegraphics[scale=0.27]{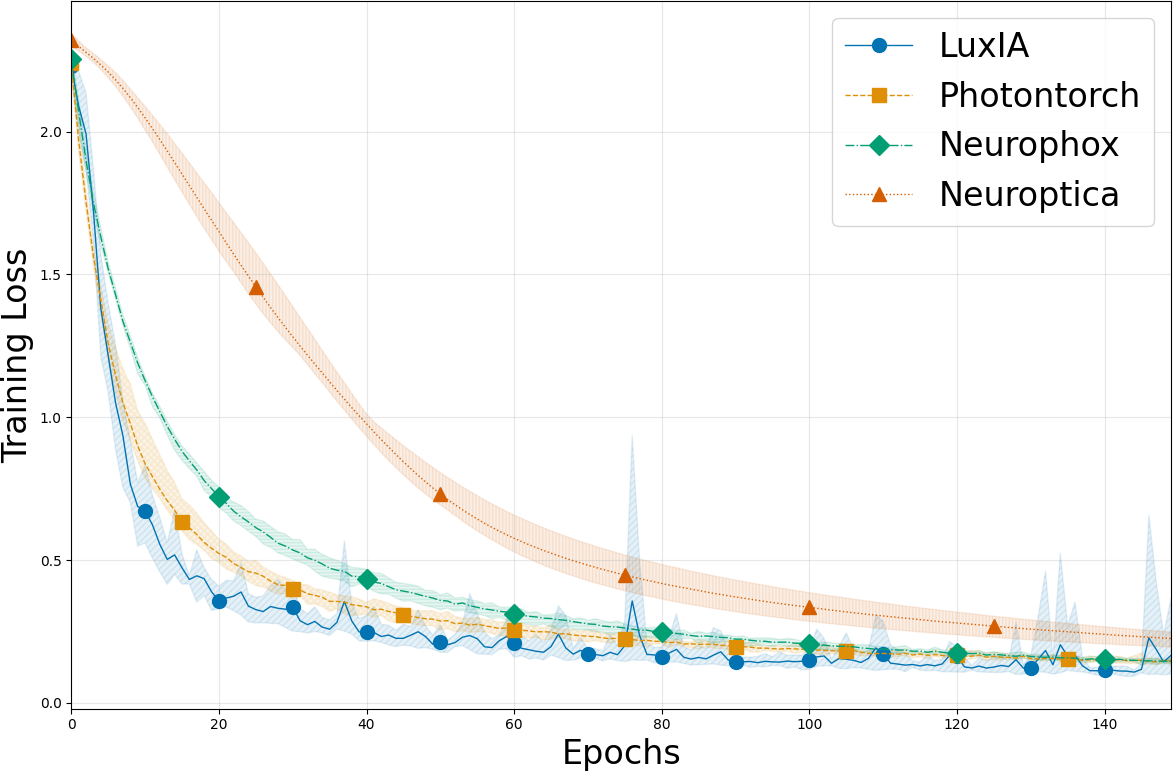}
        \label{fig:comparison_sota_tloss}
    }
    \subfigure[Validation Accuracy]{
        \centering
        \includegraphics[scale=0.27]{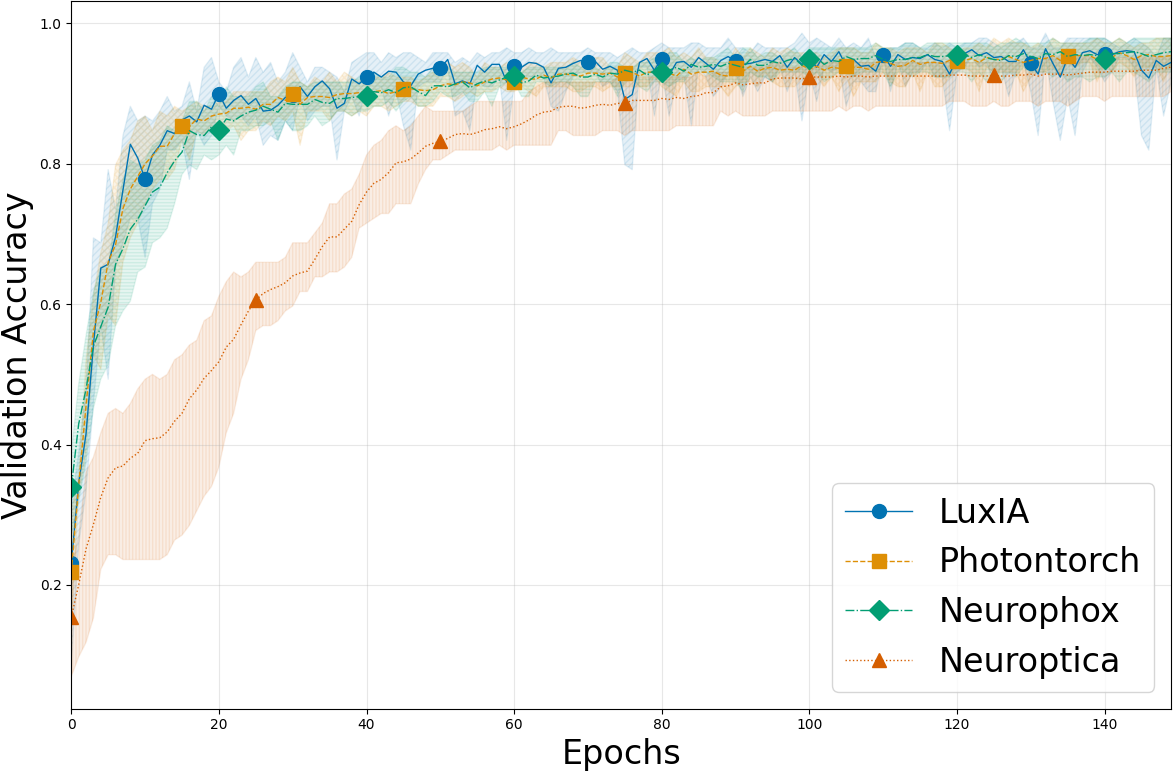}
        \label{fig:comparison_sota_vacc}
    }
    \caption{Comparison of training loss and validation accuracy with other \gls{SOTA} tools.}
    \label{fig:comparison_sota}
\end{figure*}

Neurophox, Neuroptica, and Photontorch all model the \gls{PNN} in Figure~\ref{fig:benchmark_circuit}, avoiding introducing unfairness due to modeling errors. All tools are configured with identical parameters whenever possible to ensure a proper comparison. In particular, they all employ the \gls{RMS} optimizer, with the exception of Neuroptica, which uses a custom in-situ Adam optimizer based on the adjoint method~\cite{Hughes2018}. The training is repeated 5 times for each tool, with different random seeds affecting parameters initialization and train/validation splits (the test set remains fixed). The learning rate is set to $0.0005$, and the batch size is 512. Each training is performed for 150 epochs, and the training and validation curves are plotted over these epochs. The results of the training and validation phases are presented in Figure \ref{fig:comparison_sota}, which shows the average training loss and average validation accuracy over five runs for each tool. The shaded area represents the minimum and maximum values obtained during the multiple runs.

Figure~\ref{fig:comparison_sota_tloss} shows that all tools have similar training loss trends, with LuxIA converging fastest, followed by Photontorch, Neurophox, and Neuroptica. After 150 epochs, final training losses are 0.1462 (Neurophox), 0.1470 (Photontorch), 0.2258 (Neuroptica), and 0.1667 (LuxIA). 

Figure~\ref{fig:comparison_sota_vacc} presents validation accuracy, where all tools show comparable trends. Final validation accuracies are 95.97\% (Neurophox), 94.79\% (Photontorch), 93.75\% (Neuroptica), and 94.44\% (LuxIA), confirming effective \gls{PNN} training across frameworks.

Closer analysis shows that Neuroptica's training loss decreases more slowly and its validation accuracy is less stable during the first 80 epochs, likely due to the behavior of its optimizer's settings. Since its performance later aligns with the other tools, it is worth mentioning that this optimizer might benefit from further hyperparameter search, which is beyond the scope of this comparison.

For test set evaluation, LuxIA achieves an average accuracy of 94.05\%, Neurophox 95.97\%, and Photontorch 94.79\%. Neuroptica's test accuracy is not reported, as it cannot save trained models for later evaluation. Overall, all frameworks demonstrate equivalent functionality and training dynamics. These results validate LuxIA as a reliable and practical tool for training \glspl{PNN} and confirm that the Slicing method does not introduce any significant deviation.

\subsection{Evaluation of Hardware Resource Efficiency} \label{subsec:hardware_comparison}

While the previous experiments established functional training equivalence, this section evaluates the hardware resource efficiency of all tools. The goal is to comprehensively understand the memory consumption and execution time for each tool. Two scenarios have been considered:

\begin{enumerate}
    \item \emph{Batch Size-Dependent Scenario} (\ref{subsubsec:bz_scenario}): 
        This scenario focuses on the impact of batch size on hardware resources. Increasing or decreasing the batch size can
        significantly affect memory usage and execution time. Generally, larger batch sizes lead to faster execution times
        but higher memory consumption and vice versa~\cite{Gao2021}. With a fixed \gls{PUM} mesh size ($N=800$), we observe the
        behavior of the tools with different training batch sizes, expecting to see that execution time decreases as batch size increases.
    
    \item \emph{Mesh Size-Dependent Scenario} (\ref{subsubsec:n_scenario}): 
        This scenario focuses on the impact of \gls{PUM} mesh size on memory consumption and execution time. We focus on \gls{PUM}
        mesh size for two reasons: first, when the \gls{PUM} mesh is large, most of the computational time and memory usage is
        spent computing the output of the \gls{PNN}; second, for tools relying on the transfer matrix method, the matrices
        produced by the \gls{PNN} increase quadratically with the number of inputs, as seen in Section~\ref{sec:background},
        leading to increased memory consumption and execution time, without optimizations. In this scenario, we vary
        the \gls{PUM} mesh size while keeping the batch size constant (128).
\end{enumerate}

 Measurements of time and memory usage were
meticulously recorded using Python libraries such as \texttt{time}, \texttt{psutil}, and \texttt{pynvml}. Both \gls{CPU}-only and
combined \gls{CPU}-\gls{GPU} configurations were tested, ensuring that all experiments maintained identical settings across tools.
Measurements were conducted over a single
training and validation epoch, with the test set excluded from these measurements.

\subsubsection{Batch Size-Dependent Scenario Results} \label{subsubsec:bz_scenario}
\begin{figure*}[ht] 
    \centering
    \subfigure[Memory Usage vs Training Batch Size]{
        \centering
        \includegraphics[scale=0.27]{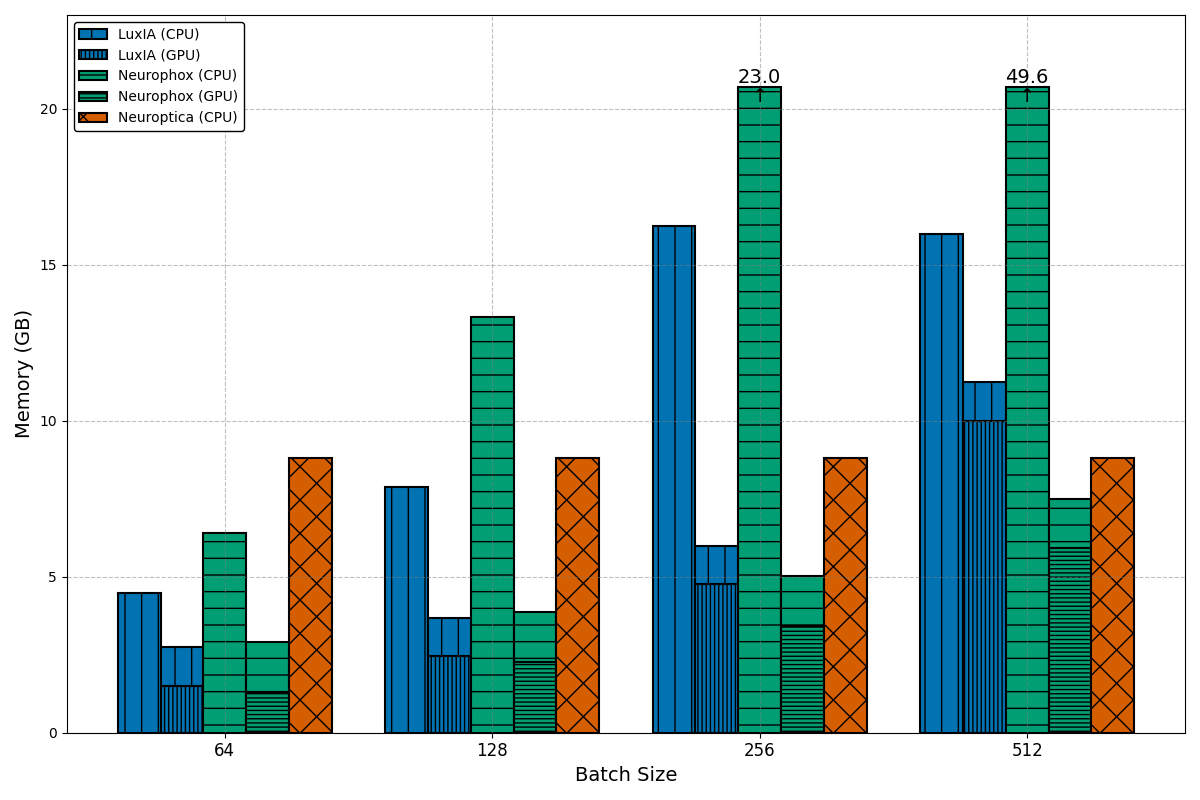} 
        \label{fig:benchmark_bs_memory} 
    }
    \subfigure[Execution Time vs Training Batch Size]{
        \centering 
        \includegraphics[scale=0.27]{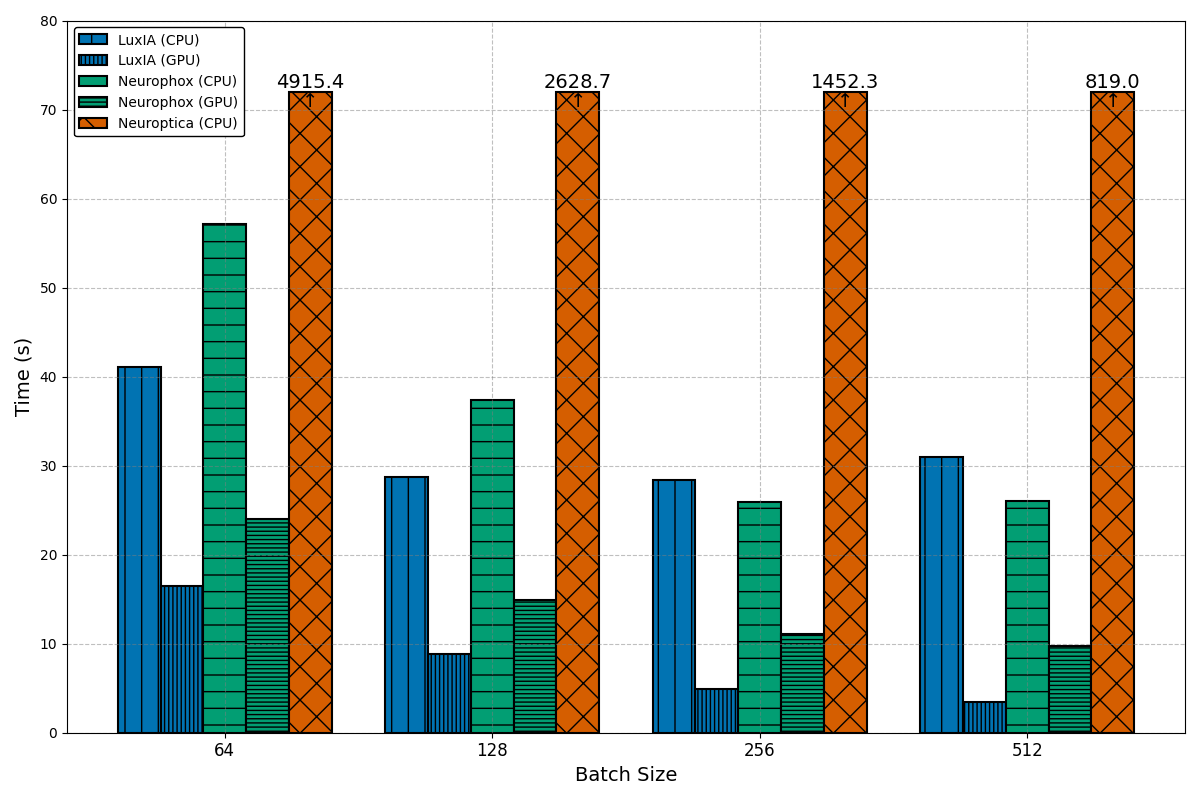}
        \label{fig:benchmark_bs_time}
    }
    \caption{Memory usage and execution time for varying training batch sizes with a fixed \gls{PUM} mesh size of $N=800$.}
    \label{fig:benchmark_bs}
\end{figure*}

In this set of experiments, the \gls{PUM} mesh size was fixed at $N=800$, and batch sizes were varied exponentially from 64 to 512.
The results are reported in Figure \ref{fig:benchmark_bs}.

LuxIA proved to be the most efficient framework in all configurations. In \gls{CPU}-only mode, it achieved competitive execution times (41.05s–30.95s) with a low memory footprint (4.47–15.99GB). With \gls{GPU} acceleration, execution times dropped from 16.45s (batch size 64) to 3.48s (batch size 512), a 4.7$\times$ speedup. LuxIA also scaled \gls{GPU} memory efficiently (1.51–9.99GB) and maintained minimal \gls{CPU} memory usage (around 1.24GB). Figures \ref{fig:benchmark_bs_memory} and~\ref{fig:benchmark_bs_time} highlight LuxIA's favorable memory scaling and execution time, especially at larger batch sizes.

Photontorch could not complete benchmarking for any batch size due to excessive memory requirements, consistently terminating with memory errors. This limitation makes it impractical for batch processing and excludes it from Figure~\ref{fig:benchmark_bs}.

Neuroptica showed moderate memory efficiency on \gls{CPU} (8.80–8.82GB across batch sizes), but at a high computational cost: execution times ranged from 4,915.36s (batch size 64) to 818.96s (batch size 512), as shown in Figures~\ref{fig:benchmark_bs_memory} and~\ref{fig:benchmark_bs_time}. Despite lower memory usage at larger batch sizes, its slow execution makes it unsuitable for time-sensitive tasks.

Neurophox ran successfully across all batch sizes and on both \gls{CPU} and \gls{GPU}. On \gls{CPU}, memory usage increased from 6.41GB (batch size 64) to 49.64GB (batch size 512), with execution times between 57.18s and 26.02s. With \gls{GPU} acceleration, execution times improved to 24.00s–9.69s, and \gls{GPU} memory scaled efficiently (1.32–5.92GB). \gls{CPU} memory remained stable at about 1.57GB across all batch sizes.

These performance metrics establish LuxIA as a highly efficient framework for \gls{PNN} simulations with large \gls{PUM} meshes, delivering superior execution speed and memory efficiency across all tested batch sizes. For large-scale applications (batch size 512), LuxIA computes approximately 235 times faster than Neuroptica and 2.8 times faster than Neurophox, while maintaining competitive memory usage. To better highlight differences among the top-performing frameworks, the y-axes in both plots have been truncated, particularly to prevent Neuroptica's much longer execution times from dominating the scale.

\subsubsection{Mesh Size-Dependent Scenario Results} \label{subsubsec:n_scenario}
\begin{figure*}[ht]
    \centering
    \subfigure[Memory Usage vs Mesh Size]{
        \centering
        \includegraphics[scale=0.27]{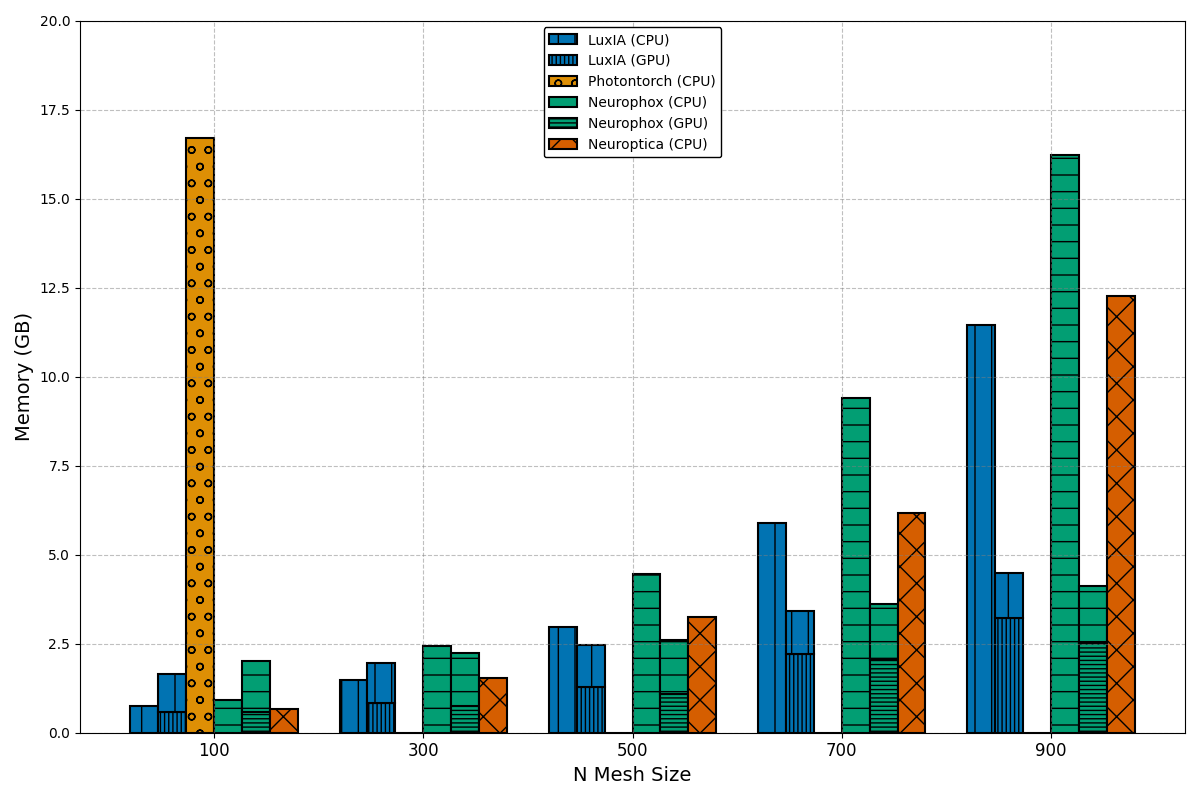}
        \label{fig:benchmark_ns_memory}
    }
    \subfigure[Execution Time vs Mesh Size]{
        \centering
        \includegraphics[scale=0.27]{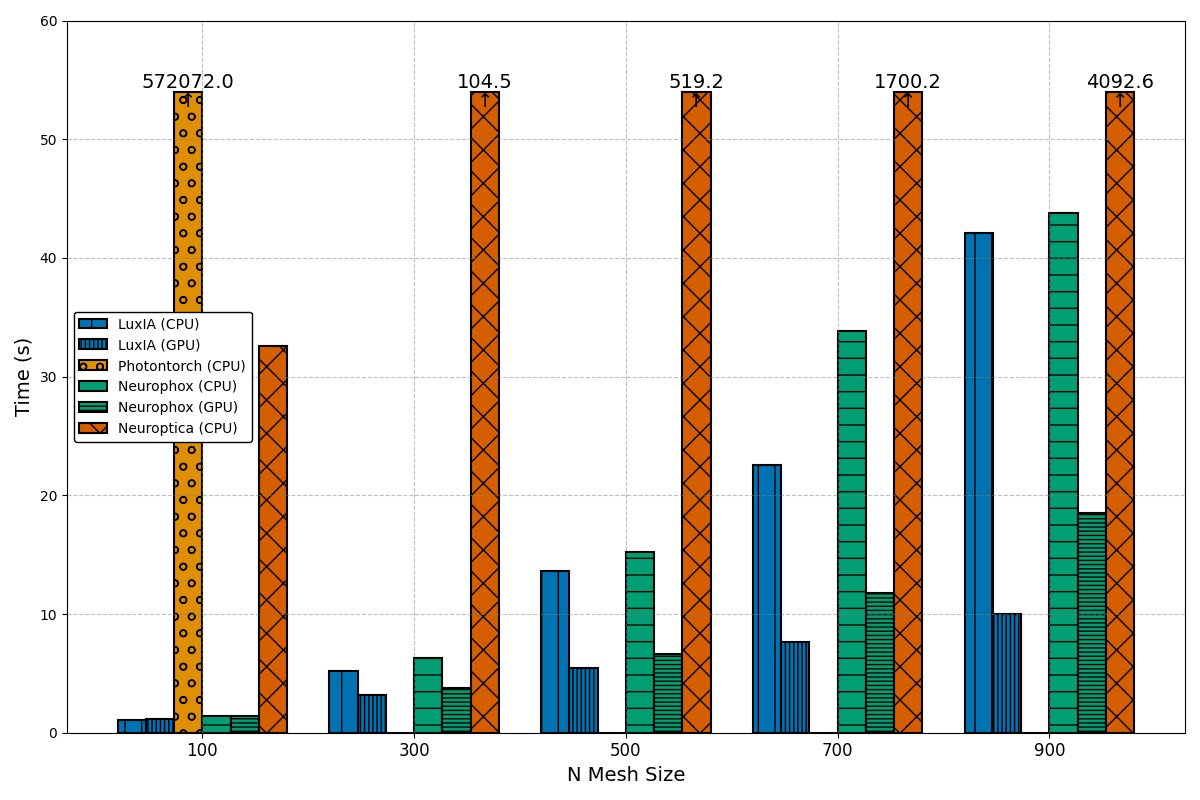}
        \label{fig:benchmark_ns_time}
    }
    \caption{Memory usage and execution time for varying \gls{PUM} mesh sizes with a fixed batch size of 128.}
    \label{fig:benchmark_ns}
\end{figure*}

In this set of experiments, the batch size was fixed at 128, and the \gls{PUM} mesh size was varied linearly from 100 to 900 in
steps of 200. The results are reported in Figure \ref{fig:benchmark_ns}.

The comparative analysis of this scenario highlights substantial differences in performance and efficiency among the tested frameworks. Across all mesh sizes and hardware configurations, LuxIA consistently outperformed the alternatives. On \gls{CPU}, LuxIA achieved the lowest execution times, ranging from 1.10s ($N=100$) to 42.09s ($N=900$), and maintained the most efficient memory usage, scaling from 0.75GB to 11.46GB. When leveraging \gls{GPU} acceleration, LuxIA further reduced execution times to a range of 1.18s ($N=100$) to 10.04s ($N=900$), representing an approximate 4.2$\times$ speedup at the largest mesh size. Its \gls{GPU} memory usage remained efficient, between 0.58GB and 3.21GB, with minimal \gls{CPU} memory overhead (1.08GB to 1.27GB). These results, visualized in Figures~\ref{fig:benchmark_ns_memory} and \ref{fig:benchmark_ns_time}, underscore how useful the \emph{Slicing method} is in execution time and memory efficiency.

In contrast, Photontorch was only able to complete the benchmarking process for the smallest mesh size ($N=100$) on
\gls{CPU}, consuming 16.69GB of memory and taking 572071.96s ($\approx$6.6 days) to execute. The rest of the benchmarking process was unable to complete due to excessive memory requirements or computational inefficiencies. This limitation prevented its
inclusion in the rest of the bar charts in Figure~\ref{fig:benchmark_ns}. This indicates severe limitations for applications
requiring larger mesh sizes, rendering Photontorch impractical for a full training and validation pipeline.

Neuroptica demonstrated efficient memory scaling on \gls{CPU}, with usage increasing from 0.66GB to 12.25GB as mesh size grew. However, this came at the expense of execution time, which increased sharply from 32.62s to 4,092.63s (see Figures~\ref{fig:benchmark_ns_memory} and \ref{fig:benchmark_ns_time}). While Neuroptica's memory efficiency was comparable to LuxIA's, its longer training time becomes increasingly problematic as the scale of the problem grows, potentially leading to substantial delays for larger mesh sizes.

Neurophox successfully executed across all mesh sizes on both \gls{CPU} and \gls{GPU}. On \gls{CPU}, memory usage ranged from 0.93GB to 16.21GB, with execution times between 1.41s and 43.78s. With \gls{GPU} acceleration, execution times improved to 1.38s--18.53s, and \gls{GPU} memory usage scaled efficiently from 0.57GB to 2.51GB. \gls{CPU} memory usage remained stable, between 1.44GB and 1.61GB, across all mesh sizes. 

These results demonstrate LuxIA's clear advantage in both speed and memory efficiency, highlighting the limitations and strengths of the other frameworks under increasing computational demands. While Neuroptica matches LuxIA in memory usage, its execution times are prohibitively long at larger meshes. Neurophox offers more balanced performance but remains less efficient than LuxIA, and Photontorch is extremely prohibitive in time and memory usage for large meshes.  

\subsection{Use cases} \label{subsec:use_cases}

After validating LuxIA's functionality and efficiency, the following experiments aim to demonstrate the framework's capability to
train different \glspl{PNN} with different \glspl{PUM} meshes and diverse datasets. Four widely-used \gls{PUM}
meshes—Clements~\cite{Clements2016}, Fldzhyan~\cite{Fldzhyan2020}, and their Bell~\cite{Bell2021} variants are evaluated against
four datasets of increasing complexity: Iris~\cite{Fisher1936}, Digits~\cite{Seewald}, MNIST~\cite{Lecun1998}, and Olivetti
Faces~\cite{Samaria1994}.

\begin{figure}[ht]
    \centering
    \includegraphics[scale=0.55]{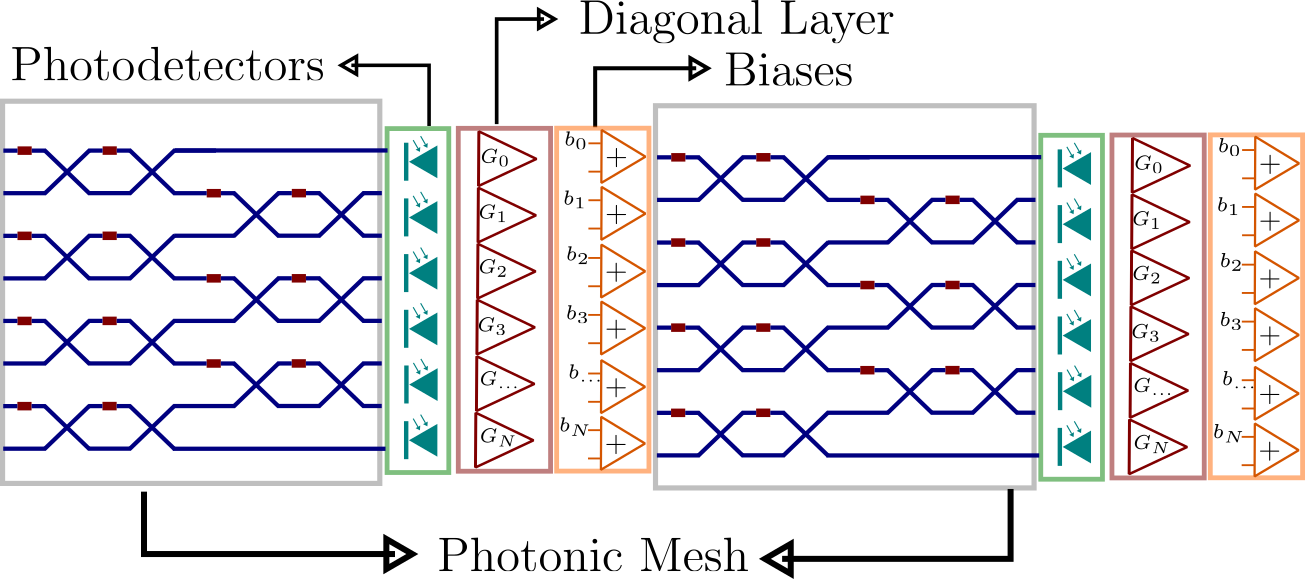}
    \caption{
    \gls{PNN} used in the experiments. The circuit consists of two \gls{PUM} meshes (Clements, Fldzhyan, or their
        Bell variants) connected in series, followed by photodetectors, a bias layer, and a diagonal layer.
    }
    \label{fig:photonic_circuit_topology}
\end{figure}

While the Digits dataset has already been described in the previous section, the other selected datasets challenge the framework in terms of memory usage and computational complexity:

\begin{itemize}
    \item \emph{Iris}: A dataset with 150 samples and 4 features, ideal for rapid prototyping and architecture testing.
    \item \emph{MNIST}: A dataset with 70,000 samples and 784 features, creating meshes of size $784 \times 784$. While
        fabrication of such large meshes remains beyond current capabilities, this dataset tests the framework's ability to handle
        computationally intensive training with large datasets. The MNIST dataset is a standard benchmark in the field. Due to the
        dataset size, if the toolchain is not optimized, it may take a long time to train the model,
        making it a good test case to evaluate the framework's performance.
    \item \emph{Olivetti}: A dataset of 400 samples with 1,024 features that has not previously been used in the context of
        \glspl{PNN}. The face classification task presents a novel challenge for photonic computing, requiring meshes of size
        $1024 \times 1024$. This dataset challenges the framework with a high feature count and complex patterns. It is an
        interesting test case for assessing its capabilities to train large \glspl{PNN} on complex tasks.
\end{itemize}

Table~\ref{tab:results_by_dataset} summarizes performance across datasets and \gls{PUM} meshes, showing final training loss, validation accuracy, and test accuracy. All experiments used the same \gls{PNN} topology (Figure~\ref{fig:photonic_circuit_topology})—two \gls{PUM} meshes, photodetectors, a bias layer, and a diagonal layer, differing only by the mesh type. This topology is the most commonly used in the literature
~\cite{Clements2016,Fldzhyan2020,Bell2021} and enables fair comparison, as differences in results reflect only mesh optimization. Each configuration was trained with the \gls{RMS} optimizer, run five times, and averaged.

\begin{table*}[ht]
    \centering
    \caption{Performance metrics by dataset and \gls{PNN}. Each entry shows final:  training loss / validation accuracy / test accuracy.}
    \label{tab:results_by_dataset}
    \resizebox{\textwidth}{!}{
        \begin{tabular}{|c|c|c|c|c|c|}
            \hline
            \textbf{Dataset (Mesh, Epochs)} & \textbf{Clements} & \textbf{Clements Bell} & \textbf{Fldzhyan} & \textbf{Fldzhyan Bell} \\
            \hline
            \textbf{Iris (4×4, 400)} 
                                            & 0.0964 / 97.50\% / 95.33\%  
                                            & 0.0902 / \textbf{98.75\%} / 94.00\% 
                                            & 0.0997 / 97.50\% / \textbf{96.66\%} 
                                            & 0.0896 / 97.50\% / 96.00\% \\
            \hline
            \textbf{Digits (64×64, 200)} 
                                            & 0.1400        / 94.03\%           / \textbf{94.055\%}
                                            & 0.1610        / \textbf{94.72\%}  / 91.77\%
                                            &\textbf{0.1129}/ 94.17\%          / 92.16\% 
                                            & 0.3053        / 90.56\%           / 92.66\%\\
            \hline
            \textbf{MNIST (784×784, 150)} 
                                            & 0.1026        / 97.03\%           / 97.17\%
                                            & 0.0851        / 96.06\%           / 97.16\%
                                            & \textbf{0.0588}/ \textbf{97.16\%} / 97.44\%
                                            & 0.0634        / 94.86\%         /  \textbf{97.60\%}  \\
            \hline
            \textbf{Olivetti (1024×1024, 400)} 
                                            & 0.3038        / 62.5\%        / 66.00\%  
                                            & 0.4915        / 64.37\%       / 64.25\%
                                            & 0.4533        / \textbf{73.12\%} / 70.00\% 
                                            & \textbf{0.2648} / 70\%        / \textbf{70.50\%} \\
            \hline
        \end{tabular}
    }
\end{table*}



\begin{figure*}[ht]
    \centering
    \subfigure[Training Loss.]{
        \centering
        \includegraphics[scale=0.27]{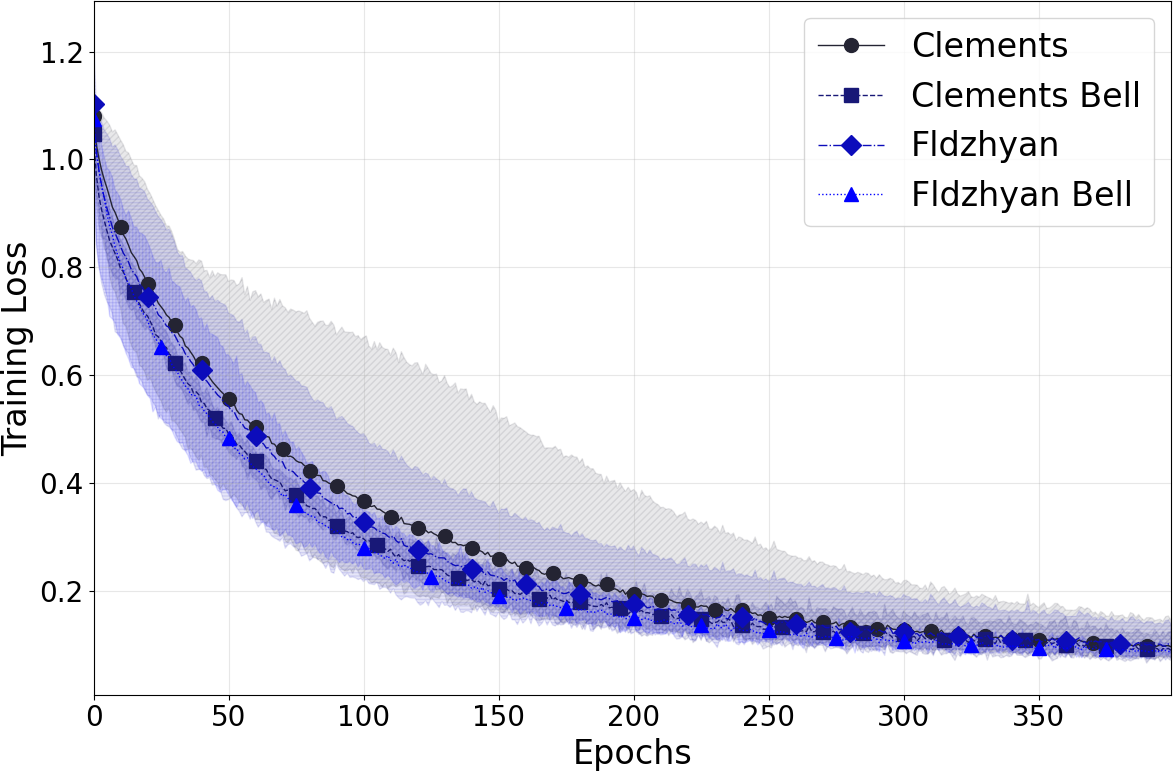}
        \label{fig:lx_iris_tloss}
    }
    \subfigure[Validation Accuracy.]{
        \centering
        \includegraphics[scale=0.27]{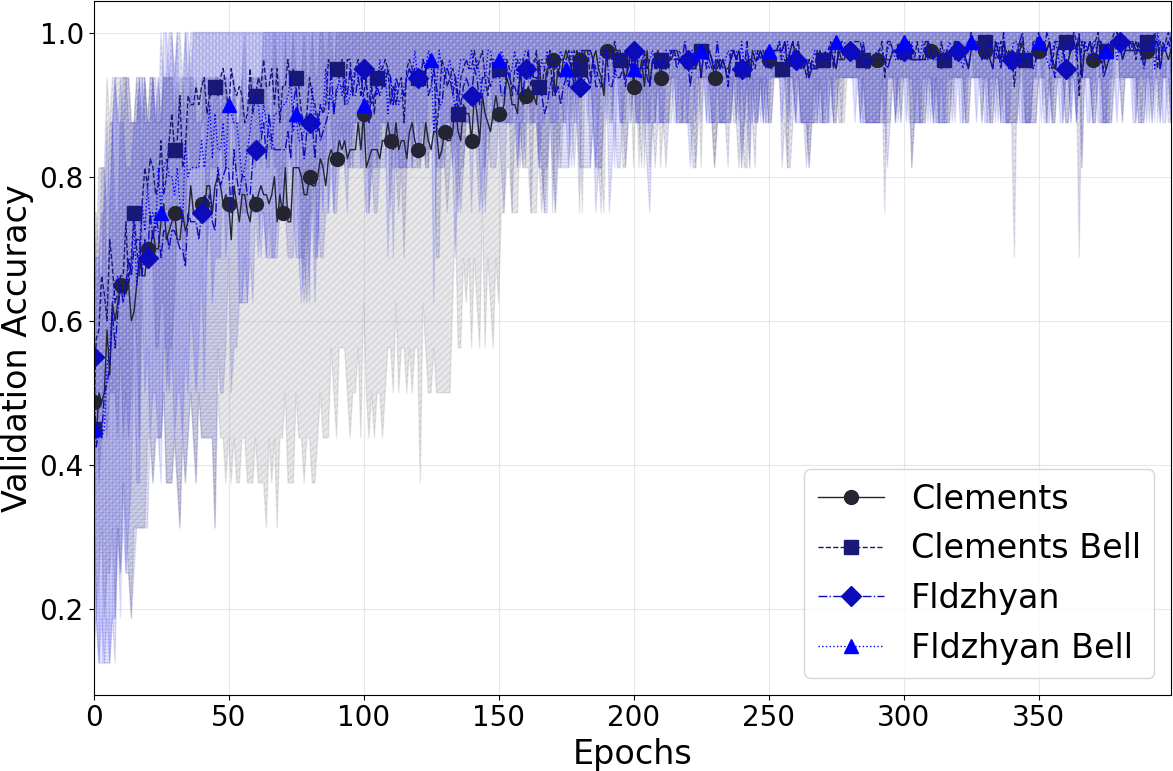}
        \label{fig:lx_iris_vacc}
    }
    \caption{Training loss and validation accuracy for the Iris dataset.}
    \label{fig:lx_iris}
\end{figure*}

All four \gls{PNN} were evaluated on the IRIS dataset over 400 epochs. As depicted in Figure~\ref{fig:lx_iris}, the models
exhibited highly effective and consistent learning behavior. The training dynamics show a rapid convergence phase within the first 100 epochs, during which training loss plummets and
validation accuracy quickly surpasses 90\%. The shaded regions in the plots, representing the standard deviation across 5 runs
with different seeds, are initially wide but narrow significantly after approximately 150 epochs. This indicates that all
\glspl{PNN} stabilize to a consistent performance trajectory. Subsequently, the models settle onto a high-performance plateau,
maintaining validation accuracies above 95\% until the end of training. This stable, high-accuracy performance suggests that all
\glspl{PNN}, no matter the \gls{PUM} mesh, effectively learned the data's underlying patterns without significant overfitting.

The final metrics, summarized in the second row in Table~\ref{tab:results_by_dataset}, confirm this strong, competitive
performance. Notably, the results highlight a subtle distinction between validation and test performance: while the \gls{PNN} with
the Clements Bell mesh obtained the highest validation accuracy of 98.75\%, the standard Fldzhyan mesh proved most effective on
unseen data, achieving the top test accuracy of 96.66\%. Meanwhile, the Fldzhyan Bell mesh reached the lowest training loss
(0.0896), suggesting the tightest fit to the training data. The close alignment between high validation and test accuracies across
all models underscores their robust generalization capabilities for this task.

\begin{figure*}[ht]
    \centering
    \subfigure[Training Loss.]{
        \centering
        \includegraphics[scale=0.27]{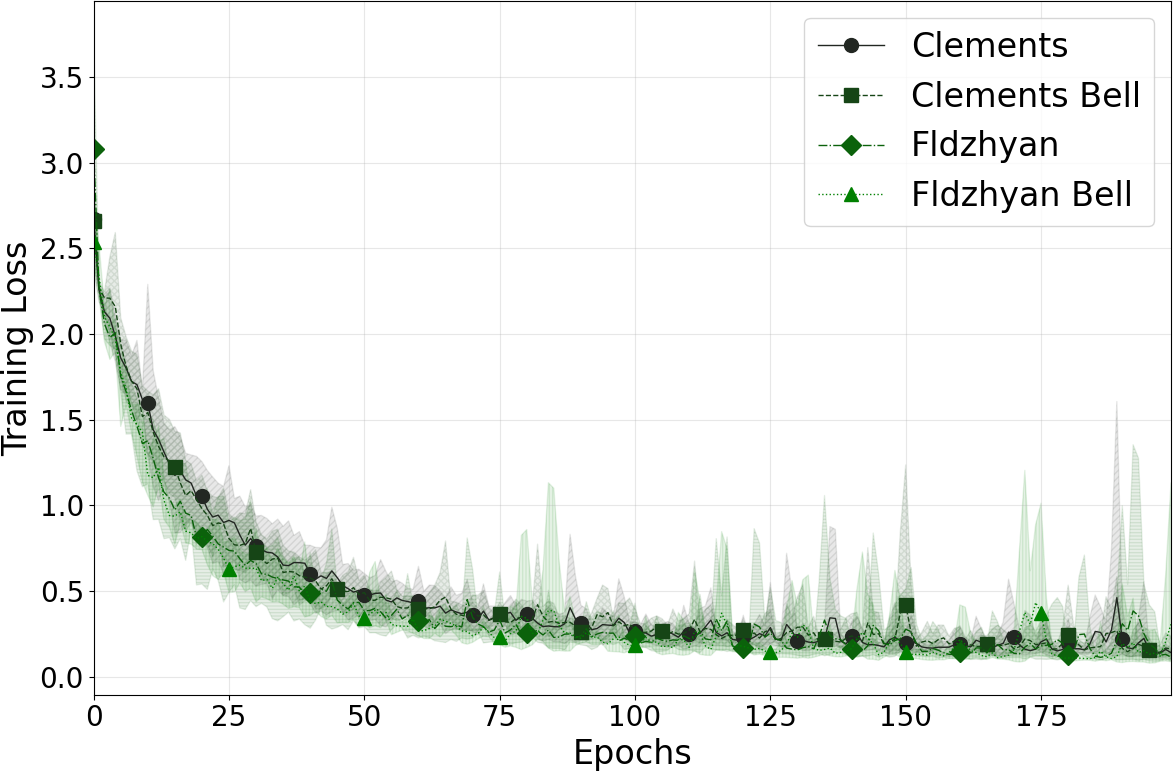}
        \label{fig:lx_digits_tloss}
    }
    \subfigure[Validation Accuracy.]{
        \centering
        \includegraphics[scale=0.27]{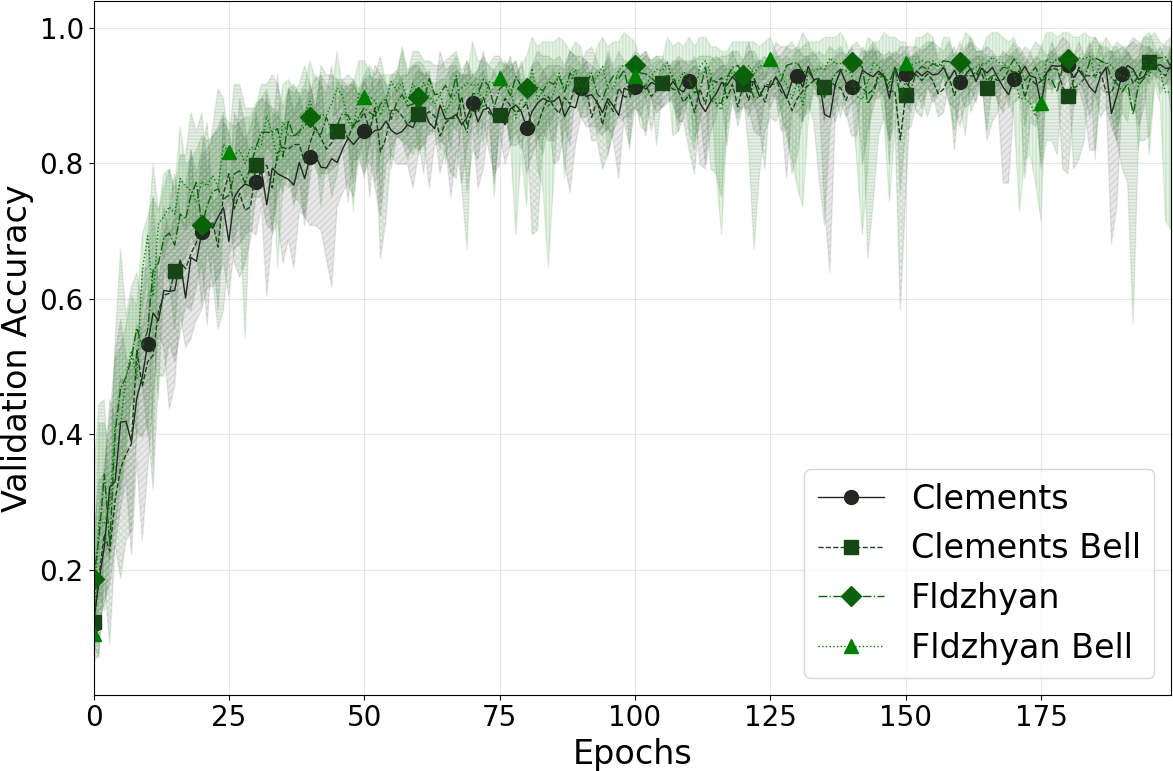}
        \label{fig:lx_digits_vacc}
    }
    \caption{Training loss and validation accuracy for the Digits dataset.}
    \label{fig:lx_digits}
\end{figure*}

Transitioning to the more complex dataset like the Digits one, the models were trained for 200 epochs, revealing significant performance distinctions
among the different \glspl{PNN}. Figure~\ref{fig:lx_digits} and
the third row of Table~\ref{tab:results_by_dataset} present the mean performance over five seeds.

The training curves in Figure~\ref{fig:lx_digits} illustrate a slight divergence that emerges within the first 50 epochs. Three
\glspl{PNN}—Clements, Clements Bell, and Fldzhyan—demonstrate robust and stable learning trajectories. In stark contrast, the
Fldzhyan Bell \glspl{PNN} struggled, exhibiting slightly wider oscillations in its training loss throughout the 200 epochs and
converging to a somewhat lower final validation accuracy plateau. This slight instability is quantified by its high final mean
training loss of 0.3053 and low mean validation accuracy of 90.56\% compared to the other \glspl{PNN}, which achieved training
losses below 0.16 and validation accuracies above 94\%.

The third row in Table~\ref{tab:results_by_dataset} reveals a nuanced trade-off between training fit, validation, and
generalization. The Fldzhyan \gls{PNN} demonstrated a superior fit to the training data, achieving the lowest final mean
training loss of 0.1129. The Clements Bell \gls{PNN} excelled on the validation set, reaching the highest mean validation accuracy of
94.72\%. However, the ultimate measure of generalization—performance on the unseen test set—was best for the Clements
\gls{PNN}, which achieved the top test accuracy of 94.06\%.

This outcome highlights that, although the meshes are theoretically equivalent for the Digits dataset, they show distinct optimization and generalization behaviors in practice. As a result, the backpropagation algorithm adjusts the parameters in a specific way for each mesh, affecting both optimization and generalization.


\begin{figure*}[ht]
    \centering
    \subfigure[Training Loss.]{
        \centering
        \includegraphics[scale=0.27]{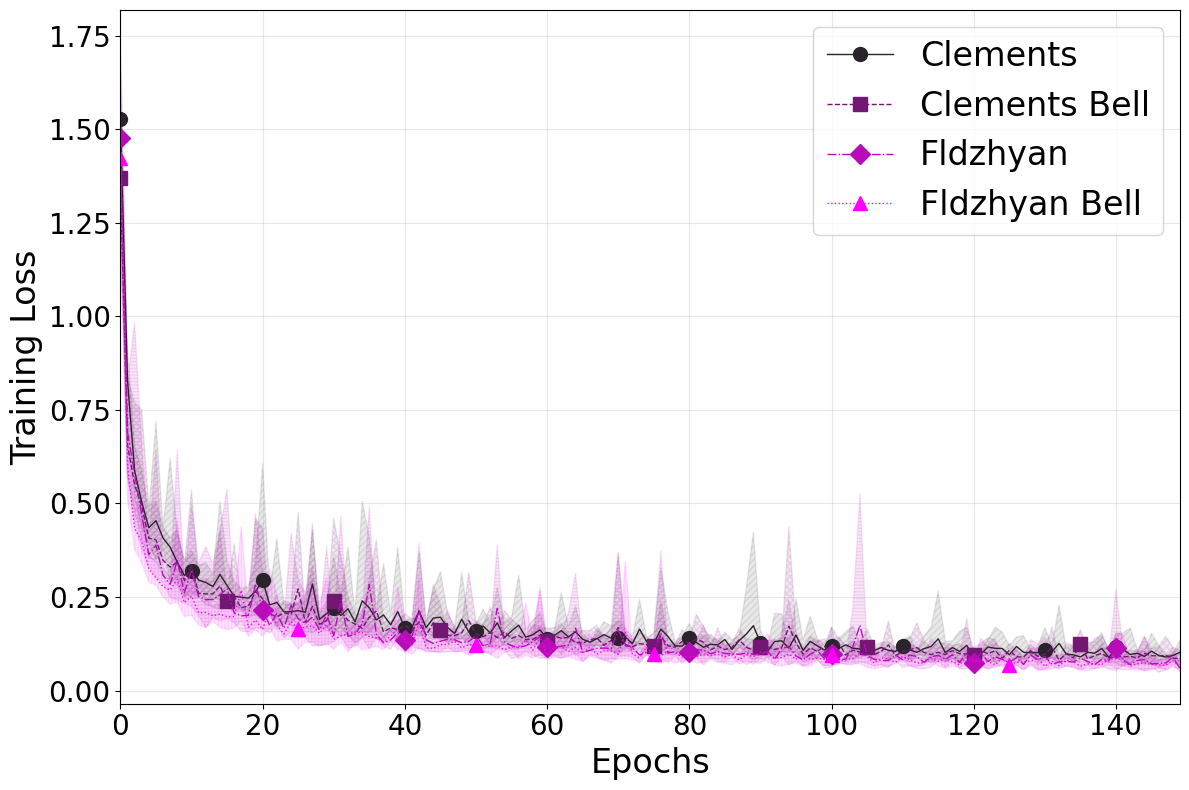}
        \label{fig:lx_mnist_tloss}
    }
    \subfigure[Validation Accuracy.]{
        \centering
        \includegraphics[scale=0.27]{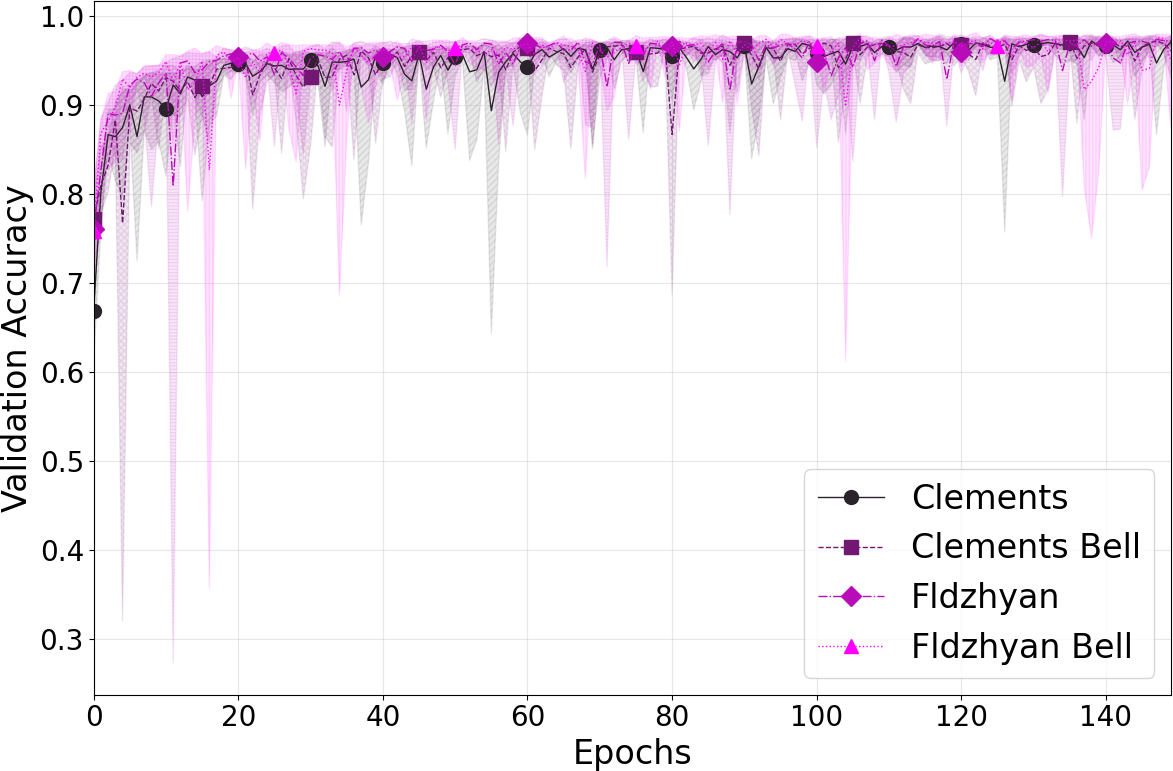}
        \label{fig:lx_mnist_vacc}
    }
    \caption{Training loss and validation accuracy for the MNIST dataset.}
    \label{fig:lx_mnist}
\end{figure*}

For the MNIST dataset, models were trained for 150 epochs using a batch size of 512 and a learning rate of 0.000038, the mean
performance over the five seeds on this task provides critical insights into the generalization capabilities of the different
\glspl{PNN}, as shown in Figure~\ref{fig:lx_mnist} and Table~\ref{tab:results_by_dataset}.

On this dataset, the learning dynamics are characterized by a rapid convergence. As depicted in Figure~\ref{fig:lx_mnist_vacc}, all
four \glspl{PNN} achieve over 90\% validation accuracy within the first 20 epochs before settling into a stable performance
plateau. The validation accuracy shows fluctuations over the five seeds, but the mean performance is over 92\% for all
\glspl{PNN} after 100 epochs. The trajectories are more stable on the training loss side, as shown in Figure~\ref{fig:lx_mnist_tloss}. Within the first 50 epochs, all models exhibit a fast decline in training loss, progressively decreasing until the
end of training. Visually, the Fldzhyan and Fldzhyan Bell models distinguish themselves by maintaining a slightly lower training
loss trajectory than the Clements-based variants, suggesting a more efficient optimization process.

At the end of training, as reported in the fourth row of Table~\ref{tab:results_by_dataset}, confirm these observations and reveal a
compelling narrative. The Fldzhyan \gls{PNN} was dominant during training and validation, securing both the lowest final
training loss (0.0588) and the highest validation accuracy (97.16\%). Based on these standard metrics, it would appear to be the
optimal model. However, the evaluation on the unseen test set presented a counterintuitive and critical result. The Fldzhyan Bell
\gls{PNN}, despite achieving the lowest validation accuracy of the group (94.86\%), delivered the highest overall test accuracy
of 97.60\%. This finding strongly suggests that the Fldzhyan Bell model developed a more robustly generalizable representation,
potentially avoiding subtle overfitting to the validation set that may have affected the other models. Furthermore, the peak test accuracy was 97.60\%. These results indicate that the framework is capable of training \glspl{PNN} effectively, obtaining similar results compared to those found in the literature, where for different \glspl{PNN}, the accuracies range from 85\% to
97\% ~\cite{Filipovich2022,Williamson2020,Roumpos2024}


\begin{figure*}[ht]
    \centering
    \subfigure[Training Loss.]{
        \centering
        \includegraphics[scale=0.27]{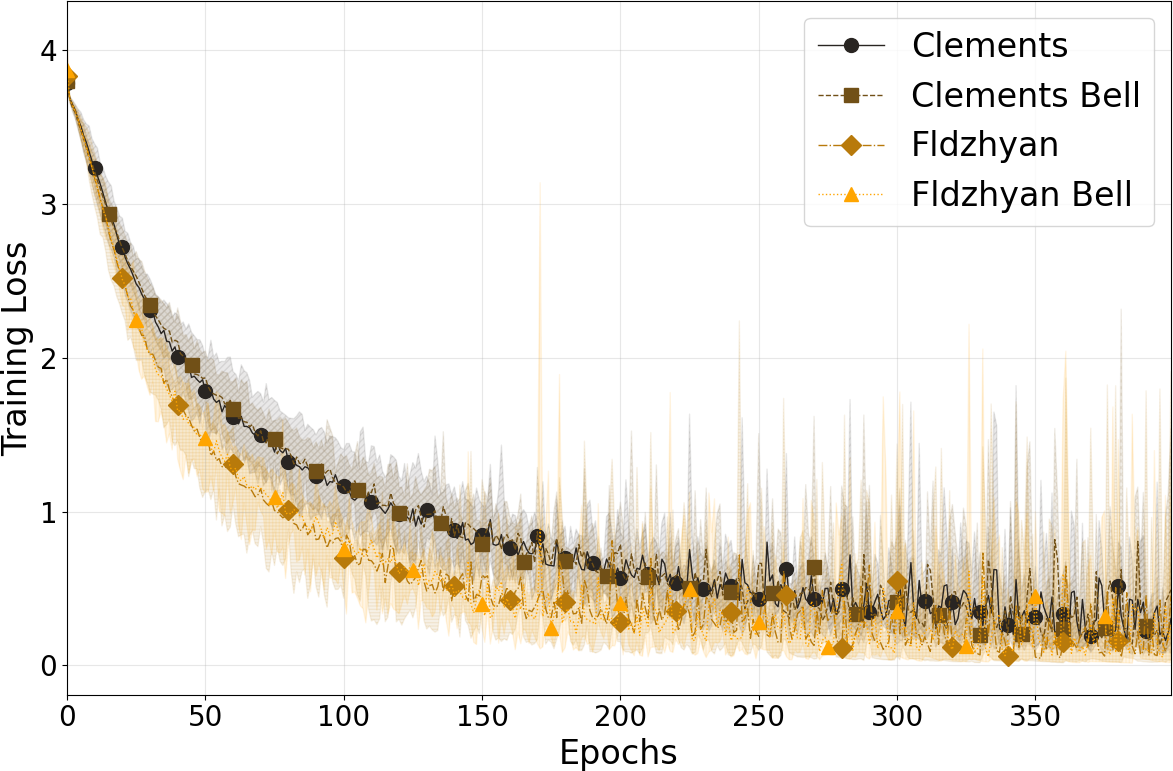}
        \label{fig:lx_olivetti_tloss}
    }
    \subfigure[Validation Accuracy.]{
        \centering
        \includegraphics[scale=0.27]{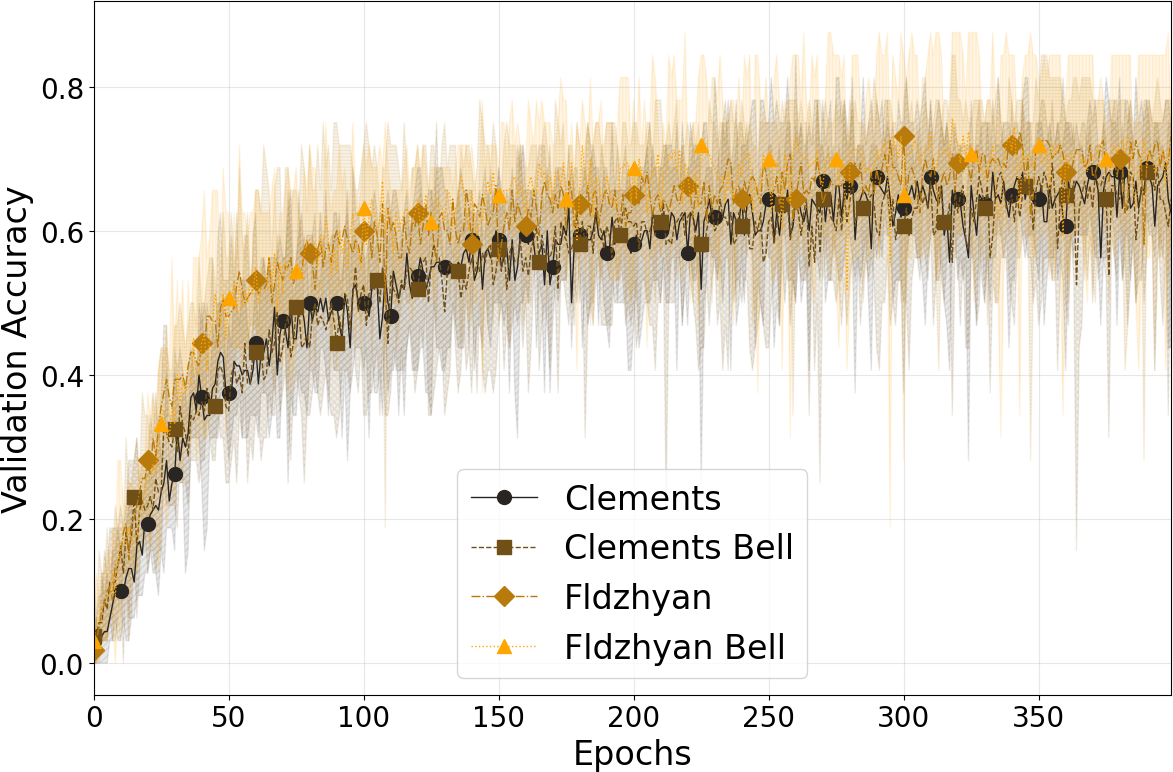}
        \label{fig:lx_olivetti_vacc}
    }
    \caption{Training loss and validation accuracy for the Olivetti Faces dataset.}
    \label{fig:lx_olivetti}
\end{figure*}

The Olivetti Faces dataset, with 400 samples and 1,024 features, constitutes the most demanding benchmark in terms of feature
dimensionality. Figure~\ref{fig:lx_olivetti} reveals significant training loss and validation accuracy instability. The
learning curves exhibit frequent oscillations and a lack of convergence, particularly for Fldzhyan \gls{PNN}. These
fluctuations suggest challenges in optimization, likely due to the small sample size relative to input complexity.

Despite these difficulties, Fldzhyan and Fldzhyan Bell achieved the highest validation accuracy (67.50\%), although both recorded
the highest training loss (0.5636). In contrast, Clements Bell yielded a lower training loss (0.4915) but slightly reduced
accuracy (64.37\%). These results, summarized in Table~\ref{tab:results_by_dataset}, highlight a trade-off between optimization
efficiency and classification performance. The performance trends in Figure~\ref{fig:lx_olivetti} suggest that further tuning or
regularization strategies may be required to achieve stable convergence on tasks involving extremely high-dimensional inputs. Such further tuning is a limitation in \glspl{PNN} rather than a simulation tool limitation.


The experimental results demonstrate that \glspl{PNN} exhibit stable learning on simpler tasks, increased convergence variability with moderate complexity, and significant optimization challenges on high-dimensional problems, with mesh architecture and sample-to-feature ratio critically influencing performance, while the framework overall proves scalable and effective across a broad spectrum of visual recognition tasks, paving the way for future advances in photonic computing.

\section{Conclusion} \label{sec:conclusion} 

The growing complexity of photonic circuits highlights key limitations in current \glspl{PNN} simulation tools, particularly regarding scalability and memory efficiency. Our evaluation shows these tools struggle with both training batch size and photonic mesh size. To address this, we introduced the Slicing Method for transfer matrix computation, enabling our LuxIA framework to simulate and train \glspl{PNN} more efficiently. The Slicing Method offers superior memory optimization, faster execution, and supports diverse architectures and datasets (e.g., MNIST, Digits, Olivetti), thus enhancing flexibility and scalability. This advancement bridges gaps in existing frameworks, allowing researchers to train larger, more complex \glspl{PNN} with lower computational costs and improved usability. The Slicing method delivers a scalable, optimized, and versatile solution for \gls{PNN} simulation, paving the way for broader adoption of photonic computing in machine learning. Future work should further refine this framework for even larger datasets and architectures.

\ifCLASSOPTIONcaptionsoff
  \newpage
\fi




\bibliographystyle{IEEEtran}
\bibliography{luxia}

%
%
%






\vspace{-20pt}
\begin{IEEEbiography}[{\includegraphics[width=1in,height=1.25in,clip,keepaspectratio]{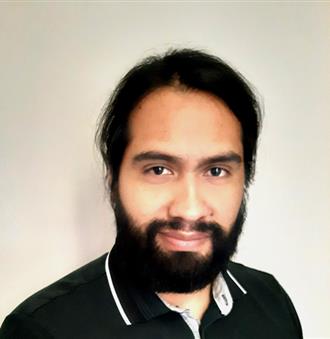}}] {Tzamn Melendez Cardosa} holds a M.S. equivalent in Electronics from Politecnico di Torino and he is a PhD student in Artificial Intelligence at Politecnico di Torino. His research focuses on the integration of photonic neural networks and RISC-V compliant interfaces to enhance system-level efficiency, with particular emphasis on the training dynamics and simulation of photonic computing architectures.
\end{IEEEbiography}
\vspace{-20pt}
\begin{IEEEbiography}[{\includegraphics[width=1in,height=1.25in,clip,keepaspectratio]{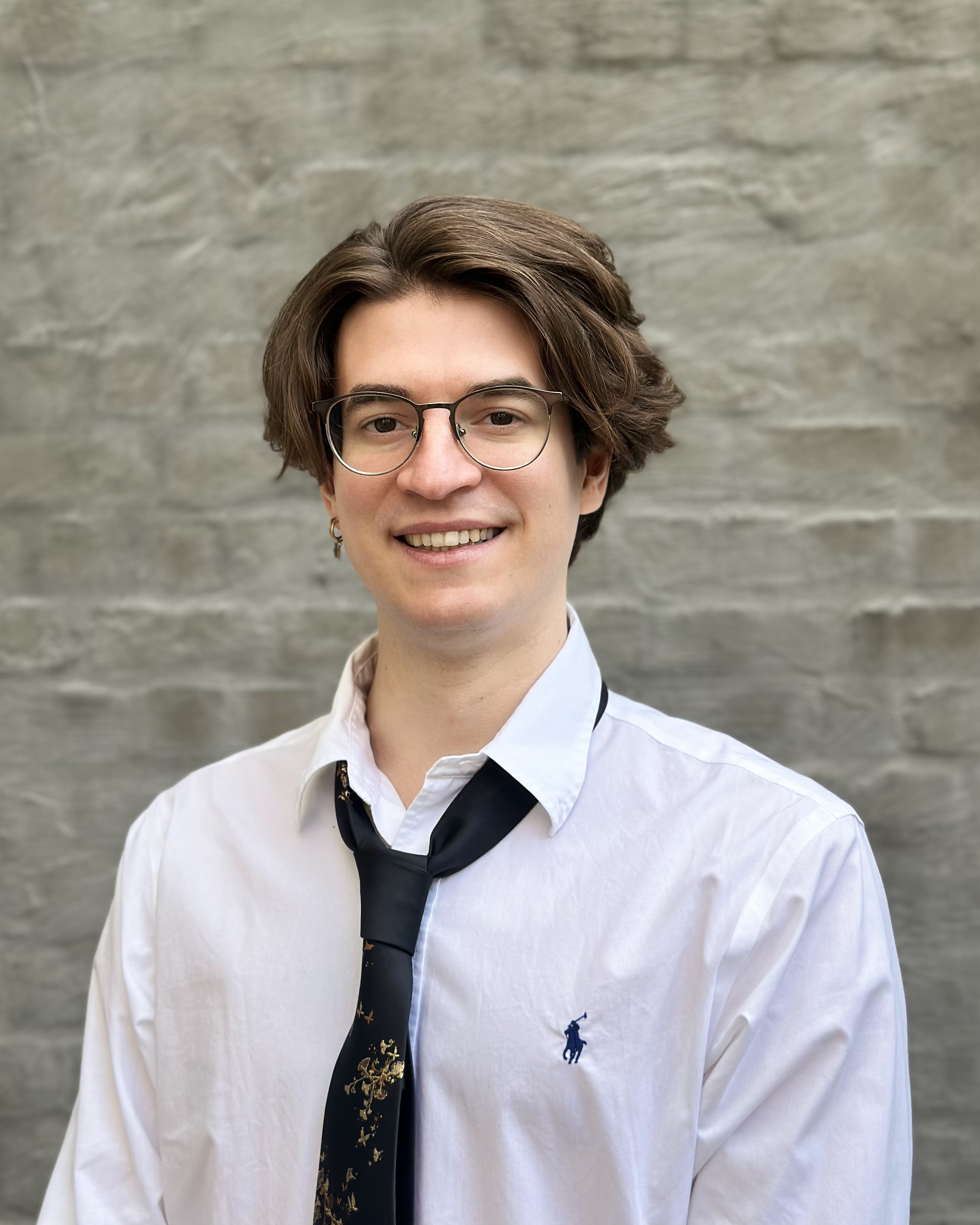}}] {Federico Marchesin} is a PhD researcher in the Photonics Research Group at Ghent University with a background in electronics from the University of Padua. His research focuses on neuromorphic photonics, specifically integrated silicon matrix-vector multiplications for neural network accelerator applications. He specializes in the simulation and modeling of photonic circuits and has significant experience in silicon integrated design.
\end{IEEEbiography}
\vspace{-20pt}
\begin{IEEEbiography} [{\includegraphics[width=1in,height=1.25in,clip,keepaspectratio]{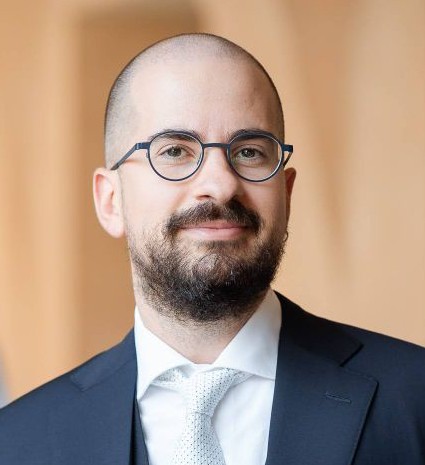}}] {Marco P. Abrate} holds a Master's degree in data science from EPFL and a Bachelor's degree in computer engineering at Politecnico di Torino, and he is a PhD candidate in neuroscience and AI at University College London. His research focuses on modelling the maturation of spatial representations in the hippocampus of developing rats using novel biologically plausible recurrent neural networks.
\end{IEEEbiography}
\vspace{-20pt}
\begin{IEEEbiography} [{\includegraphics[width=1in,height=1.25in,clip,keepaspectratio]{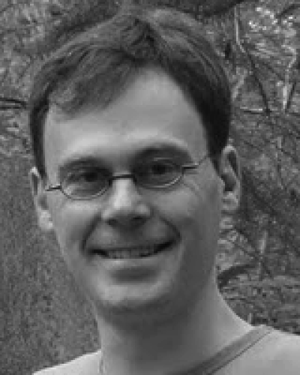}}] {Peter Bienstman} (Member, IEEE) He received the degree in electrical engineering from Ghent University, Ghent, Belgium, in 1997, and the Ph.D. degree from the Department of Information Technology (INTEC), Ghent University, in 2001 where he is currently a full professor. His research interests include several applications of nanophotonics. He has been awarded an ERC starting grant for the Naresco-project: Novel paradigms for massively parallel nanophotonic information processing.
\end{IEEEbiography}
\vspace{-20pt}
\begin{IEEEbiography}[{\includegraphics[width=1in,height=1.25in,clip,keepaspectratio]{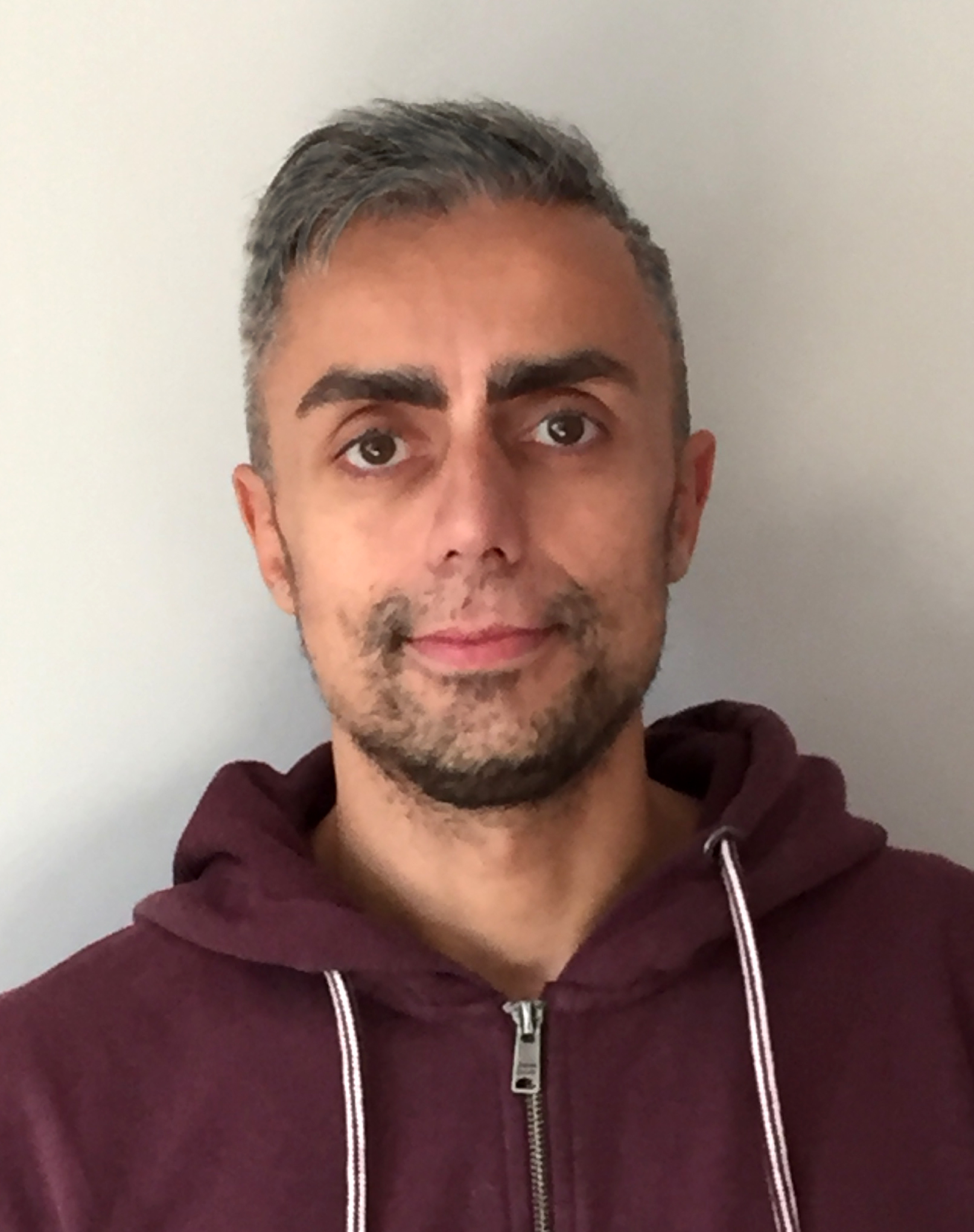}}]{Stefano Di Carlo}
(SM'00-M'03-SM'11) He received an M.Sc. degree in computer engineering and a Ph.D. in information technologies from Politecnico di Torino, Italy, where he is a Full Professor. His research interests include resilient and secure hardware architecture design, emerging computing paradigms, artificial intelligence, and neuromorphic computing. He is a Golden Core member of the IEEE Computer Society.
\end{IEEEbiography}
\vspace{-20pt}
\begin{IEEEbiography}[{\includegraphics[width=1in,height=1.25in,clip,keepaspectratio]{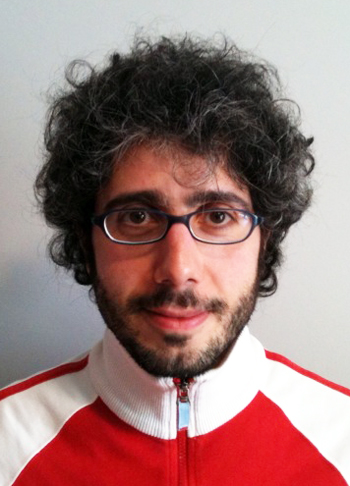}}]{Alessandro Savino}(M'14, SM'22) is an Associate Professor in the Department of Control and Computer Engineering at Politecnico di Torino (Italy). He holds a Ph.D. (2009) and an M.S. equivalent (2005) in Computer Engineering and Information Technology from the Politecnico di Torino in Italy. Dr. Savino's research contributions include Approximate Computing, Reliability Analysis, Safety-Critical Systems, Software-Based Self-Test, and Image Analysis. 
\end{IEEEbiography}

\vfill
\end{document}